\documentclass[10pt,twocolumn,letterpaper]{article}

\PassOptionsToPackage{table}{xcolor}

\usepackage[pagenumbers]{wacv} 

\usepackage[most]{tcolorbox}
\usepackage{multirow}
\usepackage{wrapfig}

\tcbset{
  colback=blue!5!white,
  colframe=blue!75!black,
  boxrule=0.8pt,
  arc=3pt,
  left=6pt,
  right=6pt,
  top=6pt,
  bottom=6pt
}
\newtcolorbox{finding}{
  colback=gray!10,
  colframe=black,
  boxrule=0.5pt
}
\usepackage{bbold}

\newcommand{\myparagraph}[1]{\smallskip\noindent\textbf{#1}}
\newcommand\blfootnote[1]{%
  \begingroup
  \renewcommand\thefootnote{}\footnote{#1}%
  \addtocounter{footnote}{-1}%
  \endgroup
}

\definecolor{wacvblue}{rgb}{0.21,0.49,0.74}
\usepackage[pagebackref,breaklinks,colorlinks,allcolors=wacvblue]{hyperref}

\newcommand{\up}[1]{\textcolor{ForestGreen}{(+#1)}}
\newcommand{\dn}[1]{\textcolor{red}{($-$#1)}}
\newcommand{\eqd}[1]{\textcolor{gray}{(#1)}}

\def\wacvPaperID{1429} 
\def\confName{WACV}
\def\confYear{2027}

\title{Probing and Leveraging Video Diffusion Transformer Features \\for Robust Point Tracking}

\author{
    Soowon Son\textsuperscript{\rm 1} \quad
    Honggyu An\textsuperscript{\rm 1} \quad
    Jisu Nam\textsuperscript{\rm 1} \quad
    Hyunah Ko\textsuperscript{\rm 1} \quad
    Chaehyun Kim\textsuperscript{\rm 1} \quad \\
    Dahyun Chung\textsuperscript{\rm 1} \quad
    Jung Yi\textsuperscript{\rm 1} \quad
    Siyoon Jin\textsuperscript{\rm 1} \quad
    Junhwa Hur\textsuperscript{\rm 2}$^{\dagger}$ \quad
    Seungryong Kim\textsuperscript{\rm 1}$^{\dagger}$ \\[7pt]
    \textsuperscript{\rm 1}KAIST AI \qquad \textsuperscript{\rm 2}Google DeepMind \\[5pt]
{\tt \href{https://cvlab-kaist.github.io/DiTracker}{\textcolor{purple}{https://cvlab-kaist.github.io/DiTracker}}}
}
\begin{document}
\maketitle

\begin{abstract}
Despite achieving strong results on standard benchmarks, current point tracking methods rely on feature backbones that are rarely designed with the temporal coherence needed for robust real-world performance. While recent works incorporate powerful visual foundation model (VFM) features into tracking pipelines, no prior work has systematically analyzed which VFM provides the most robust representations for point tracking. We present the first such analysis, evaluating diverse VFMs in a zero-shot setting on both standard and robustness benchmarks for point tracking. Our study reveals that video diffusion transformers (DiTs) consistently yield the most temporally coherent and discriminative features, even surpassing ResNet backbones explicitly supervised on tracking data. We hypothesize this advantage stem from large-scale video pretraining, full 3D spatio-temporal attention, and a diffusion training objective. Motivated by this finding, we propose \textbf{DiTracker}, which integrates video DiT features into existing tracking frameworks through query-key matching cost computation, cost-level fusion with a lightweight ResNet branch, and LoRA adaptation. Under the same tracking head, DiTracker is trained solely on synthetic data with far fewer iterations, yet outperforms CoTracker3 trained with additional real-world videos, with the largest gains under challenging and corrupted scenarios. It further generalizes across tracking heads and scales with backbone size, confirming that generative video pretraining provides real-world priors that reduce the dependence on large-scale real-data supervision.
\end{abstract}

\blfootnote{$^\dagger$Co-corresponding.}

\section{Introduction}

Point tracking~\cite{doersch2023tapir, doersch2022tap, cho2024local, karaev2024cotracker3, harley2025alltracker, zholus2025tapnext, xiao2024spatialtracker, cho2025seurat, kim2025exploring} aims to track consistent physical points across video frames, and serves as a fundamental building block for 4D reconstruction~\cite{wang2024shape, feng2025st4rtrack, chen2025back}, robotics~\cite{vecerik2024robotap}, and video editing~\cite{geng2024motionprompting}. Most existing approaches follow a common paradigm~\cite{doersch2023tapir, cho2024local, karaev2024cotracker3, harley2025alltracker, aydemir2025track, xiao2024spatialtracker}, where a feature backbone extracts per-frame representations, matching costs are computed from these features, and a tracking head refines trajectories conditioned on both. While this design captures inter-frame correspondences effectively, the quality of the underlying feature representations remains largely underexplored.

Recent advances in point tracking~\cite{cho2024local, karaev2024cotracker3, harley2025alltracker, zholus2025tapnext} have focused mainly on improving the tracking head, while relying on temporally independent backbones (\eg, ResNet~\cite{he2016deep}) that lack temporal consistency across frames. As a result, these methods usually struggle under real-world challenges such as motion blur, fast motion, and frequent occlusions. A few works~\cite{aydemir2025track, lai2026cowtracker} have begun replacing conventional 2D CNN backbones with Visual Foundation Models (VFMs) such as DINOv2~\cite{oquab2023dinov2}. However, because foundation models differ drastically in their architectures and training objectives, which VFM offers the most suitable representation for dense temporal tracking remains unclear. This raises a fundamental question: \textbf{which VFM provides the most robust feature representation for point tracking, and how can we best leverage it?}

We answer this question through the first systematic study of VFM representations for robust point tracking. We benchmark a diverse set of models spanning pretraining modalities (\eg, image or video), architectures (\eg, UNet or Transformer), and learning objectives (\eg, MAE or diffusion). To rigorously assess robustness, we augment TAP-Vid-DAVIS~\cite{doersch2022tap} with motion blur perturbations~\cite{hendrycks2019benchmarking} and evaluate on ITTO-MOSE~\cite{demler2025tracker}, which features diverse motion dynamics and frequent occlusions. Our analysis reveals a clear trend: \textbf{video diffusion transformers (DiTs)~\cite{yang2024cogvideox, li2024hunyuan, wan2025wan} consistently outperform all other VFMs~\cite{assran2025v, carreira2024scaling, oquab2023dinov2, simeoni2025dinov3, esser2024scaling, blattmann2023stable} under both standard and challenging scenarios.} We attribute this advantage to three characteristics of video DiTs, namely large-scale real video pretraining, full 3D spatio-temporal attention, and a diffusion training objective.

To investigate how these features behave under supervised training, we build \textbf{DiTracker}, a simple framework that integrates video DiT features into existing tracking pipelines through three components: query-key matching cost computation, cost-level fusion with a lightweight ResNet branch that compensates for the DiT's lower spatial resolution, and LoRA adaptation that transfers the pretrained features to tracking.

We evaluate DiTracker on both standard and robustness benchmarks used in our feature analysis. Trained solely on synthetic data, either Kubric~\cite{greff2022kubric} alone or augmented with fewer than 1k additional synthetic samples, DiTracker outperforms CoTracker3 trained with additional real-world videos under the same tracking head, with particularly large gains under corruptions. Moreover, this benefit is not tied to a single design, as video DiT features generalize across different tracking heads and scale with backbone size. These results show that, with the same tracking head, video DiT features can replace large-scale real-data supervision.

In summary, our contributions are as follows:
\begin{itemize}
  \renewcommand{\labelitemi}{$\bullet$}
    \item We present the first systematic analysis of VFM features for robust point tracking, revealing video DiTs as the most robust backbone, even surpassing ResNet backbones supervised on tracking data.
    \item We introduce \textbf{DiTracker} to leverage video DiT features for supervised tracking, identifying LoRA adaptation, query-key matching, and cost-level fusion as its key designs.
    \item With only synthetic data and far fewer iterations, DiTracker outperforms its same-head baseline CoTracker3 trained on real-world videos, with large gains under corruptions, and generalizes across tracking heads and backbone sizes.
\end{itemize}

\section{Related Work}

\myparagraph{Feature backbones for point tracking.}
While recent progress has focused on sophisticated refinement modules~\cite{doersch2023tapir, cho2024local, karaev2024cotracker, doersch2022tap}, most methods still depend on standard backbones such as ResNet~\cite{he2016deep} or TSM-ResNet~\cite{lin2019tsm}, leaving the potential of stronger feature extractors underexplored.
To address this, several works have investigated DINOv2~\cite{oquab2023dinov2} as a backbone, demonstrating substantial gains for point tracking~\cite{kim2025exploring, aydemir2025track, tumanyan2024dino}. More broadly, \cite{aydemir2024can} showed that representations from diverse vision foundation models can further improve tracking quality, with Stable Diffusion~\cite{rombach2022high} features even surpassing those of DINOv2. In the same direction, DiffTrack~\cite{nam2025emergent} revealed that video diffusion transformers (DiTs), even without explicit supervision for the tracking task, possess layers highly aligned with temporal correspondence and outperform conventional backbones in zero-shot settings. However, there has been no work that systematically analyzes the robustness of VFMs for point tracking. Our work exploits video DiTs as robust feature backbones and integrates their representations seamlessly into existing point tracking frameworks.

\myparagraph{Diffusion models for geometric tasks.}
Beyond generation, large-scale pre-training endows diffusion models with rich representations~\cite{ho2020denoising, rombach2022high} applicable to geometric tasks such as visual correspondence~\cite{tang2023emergent, zhang2023tale, meng2024not, gan2025unleashing, nam2023diffusion}, segmentation~\cite{xu2023open}, and depth estimation~\cite{ke2024repurposing}. Pioneering approaches such as DIFT~\cite{tang2023emergent} and SD-DINO~\cite{zhang2023tale} demonstrated that diffusion models inherently encode semantic and geometry aware features, achieving competitive zero-shot correspondence performance. Subsequent studies have further strengthened these capabilities through architectural modifications~\cite{luo2023diffusion, zhang2024telling, xue2025matcha, liu2025mind}, distillation~\cite{stracke2025cleandift}, and prompt tuning~\cite{li2024sd4match}. 
This research suggests diffusion models can generalize to perception tasks with minimal supervision with synthetic dataset~\cite{ke2024repurposing}, reducing the sim-to-real gap that limits conventional point tracking backbones.
We extend this insight by repurposing video diffusion features to achieve robust tracking using only sparse, high-quality synthetic data.

\begin{table}
    \centering
    \footnotesize
    \caption{\textbf{Details of the Visual Foundation Models (VFMs) probed for point tracking.} For each VFM, we report the parameter count, the scale of its image and video training data, the backbone architecture, and the training objective.}
    \resizebox{\linewidth}{!}{
    \begin{tabular}{c|c|c|c|c|c}
        \toprule
        \multirow{2}{*}{Method}& \multirow{2}{*}{Parameter} & \multicolumn{2}{c|}{Training Data}& \multirow{2}{*}{Architecture}& \multirow{2}{*}{Objective}\\
        &  &Image&Video & & \\
        \midrule
        DINOv2-B/14& 86M &LVD-142M (142M)&- & Transformer & Self-Sup. \\
        DINOv3-B/16& 86M &LVD-1689M (1.7B)&- & Transformer & Self-Sup. \\
        DINOv3-7B/16& 7B &LVD-1689M (1.7B)&- & Transformer & Self-Sup. \\
        SD3 (Medium)& 2B &Unknown(1B)&- & Transformer & Diffusion \\
        SVD& 1.5B &SD2.1 Data (2B)&LVD-F (152M) & UNet & Diffusion \\
        V-JEPA2-G/16& 1B &ImageNet (1M) &VideoMix22M (1M hours)& Transformer & MAE \\
        HunyuanVideo& 13B &Unknown (Internet Scale)&Unknown (Internet Scale) & Transformer & Diffusion \\
        WAN-1.3B& 1.3B &Unknown (10B)&Unknown (1.5B) & Transformer & Diffusion \\
        WAN-14B& 14B& Unknown (10B)& Unknown (1.5B)& Transformer &Diffusion \\
        CogVideoX-2B& 2B& Unknown (2B)& Unknown (35M)& Transformer &Diffusion \\
        CogVideoX-5B& 5B& Unknown (2B)& Unknown (35M)& Transformer &Diffusion \\
        \bottomrule
    \end{tabular}
    }
    \label{tab:vfm_details}
\end{table}
\section{Probing VFMs for Point Tracking}
\label{sec:analysis}
We systematically analyze the zero-shot point tracking performance of various Visual Foundation Models (VFMs) across different pretraining modalities (\eg, image or video), architectural designs (\eg, UNet or Transformer), and learning objectives (\eg, MAE or Diffusion), under both standard and challenging tracking scenarios.

\subsection{Analysis setup}

\myparagraph{Comparisons.}
We include image-trained models DINOv2~\cite{oquab2023dinov2} and DINOv3~\cite{simeoni2025dinov3}, which are self-supervised vision transformers, and Stable Diffusion 3 (SD3)~\cite{esser2024scaling}, an image diffusion transformer (DiT). For video-trained models, we include V-JEPA2~\cite{assran2025v}, which is transformer models trained with MAE objectives. SVD~\cite{blattmann2023stable} is a U-Net-based video diffusion model. Finally, HunyuanVideo~\cite{kong2024hunyuanvideo}, CogVideoX~\cite{yang2024cogvideox}, and WAN2.1~\cite{wan2025wan} are video DiTs with full 3D spatio-temporal attention. Details on those models are summarized in \Cref{tab:vfm_details}.

We also compare with the ResNet~\cite{he2016deep} backbones of point tracking models. Unlike CoTracker3~\cite{karaev2024cotracker3}, the TAP-Net~\cite{doersch2022tap}, TAPIR~\cite{doersch2023tapir}, and BootsTAPIR~\cite{doersch2024bootstap} models are trained with a global cost that resembles zero-shot point tracking. BootsTAPIR and CoTracker3 are trained on large-scale real-world video data. VGGT~\cite{wang2025vggt}, which uses DINOv2~\cite{oquab2023dinov2} as its feature backbone, adopts a 3D full-attention architecture and is trained on multiple geometric tasks including point tracking.

\myparagraph{Evaluation dataset.}
We benchmark each method across three settings: TAP-Vid-DAVIS~\cite{doersch2022tap}, TAP-Vid-DAVIS corrupted with motion blur, and ITTO-MOSE~\cite{demler2025tracker}. TAP-Vid-DAVIS is a standard point tracking benchmark with relatively simple motions. To assess robustness to visual degradation from fast motion, we corrupt it with motion blur at varying severity levels from ImageNet-C~\cite{hendrycks2019benchmarking}. ITTO-MOSE is a challenging real-world benchmark with diverse motion dynamics and frequent occlusions. We categorize its trajectories by motion dynamics (static, moderate, fast) and reappearance frequency (none, occasional, frequent), enabling analysis across difficulty levels.

\begin{table*}[t]
    \centering
\caption{\textbf{Zero-shot comparison of VFMs and point-tracking feature backbones} on TAP-Vid-DAVIS~\cite{doersch2022tap} (clean and ImageNet-C~\cite{hendrycks2019benchmarking} motion-blur corruption) and ITTO-MOSE~\cite{demler2025tracker}, all at the same $30 \times 52$ feature resolution. \textcolor{gray}{Gray} rows incidate feature backbones taken from models trained with task-specific supervision (\eg, point tracking). \textbf{Bold} and \underline{underline} denote the best and second-best results.}
    \label{tab:davis_zeroshot}
    \resizebox{0.9\textwidth}{!}{
    \begin{tabular}{l|c|ccc|c|ccc|ccc}
         \toprule
         \multirow{3}{*}{Method}&\multicolumn{4}{c|}{TAP-Vid-DAVIS}& \multicolumn{7}{c}{ITTO-MOSE}\\
         \cmidrule{2-12}
 & \multirow{2}{*}{Original}&\multicolumn{3}{c|}{\textit{Motion Blur}}& \multirow{2}{*}{Avg.}&\multicolumn{3}{c|}{\textit{Motion Dynamics}}&\multicolumn{3}{c}{\textit{Reappearance Freq.}}\\
   && 1& 3&5& & Static& Moderate& Fast& None& Occasional& Frequent\\ 
        \midrule \midrule
        \textcolor{gray}{CoTracker3~\cite{karaev2024cotracker3}}&\textcolor{gray}{35.6}& \textcolor{gray}{35.0}& \textcolor{gray}{30.1}&\textcolor{gray}{24.6}& \textcolor{gray}{23.7}& \textcolor{gray}{33.3}& \textcolor{gray}{24.2}& \textcolor{gray}{14.9}& \textcolor{gray}{32.7}& \textcolor{gray}{24.1}&\textcolor{gray}{14.6}\\
 \textcolor{gray}{TAP-Net~\cite{doersch2022tap}}& \textcolor{gray}{41.8}& \textcolor{gray}{41.0}& \textcolor{gray}{40.3}& \textcolor{gray}{38.5}& \textcolor{gray}{32.0}& \textcolor{gray}{42.0}& \textcolor{gray}{33.2}& \textcolor{gray}{22.5}& \textcolor{gray}{43.5}& \textcolor{gray}{31.5}&\textcolor{gray}{21.5}\\
 \textcolor{gray}{TAPIR~\cite{doersch2023tapir}}& \textcolor{gray}{44.3}& \textcolor{gray}{42.7}& \textcolor{gray}{41.8}& \textcolor{gray}{\textbf{40.1}}& \textcolor{gray}{32.5}& \textcolor{gray}{42.4}& \textcolor{gray}{34.5}& \textcolor{gray}{22.1}& \textcolor{gray}{43.0}& \textcolor{gray}{32.9}&\textcolor{gray}{21.8}\\
 \textcolor{gray}{BootsTAPIR~\cite{doersch2024bootstap}}& \textcolor{gray}{\underline{48.3}}& \textcolor{gray}{43.6}& \textcolor{gray}{41.5}& \textcolor{gray}{39.0}& \textcolor{gray}{\underline{34.9}}& \textcolor{gray}{44.9}& \textcolor{gray}{\underline{38.4}}& \textcolor{gray}{\textbf{23.6}}& \textcolor{gray}{\underline{46.2}}& \textcolor{gray}{\underline{35.2}}&\textcolor{gray}{\underline{23.7}}\\
 \textcolor{gray}{VGGT~\cite{wang2025vggt}}& \textcolor{gray}{44.7}& \textcolor{gray}{42.9}& \textcolor{gray}{39.6}& \textcolor{gray}{28.6}& \textcolor{gray}{29.6}& \textcolor{gray}{36.2}& \textcolor{gray}{31.0}& \textcolor{gray}{22.7}& \textcolor{gray}{37.5}& \textcolor{gray}{29.7}&\textcolor{gray}{21.9}\\
        \midrule
        DINOv2-B/14~\cite{oquab2023dinov2}    &40.5& 39.2& 34.4&29.5& 22.6& 25.0& 23.8& 19.1& 24.3& 24.3&19.4\\
        DINOv3-B/16~\cite{simeoni2025dinov3}    &41.3& 40.6& 37.5&32.2& 24.9& 30.8& 24.3& 20.3& 29.0& 26.3&19.6\\
        DINOv3-7B/16~\cite{simeoni2025dinov3}    &42.6& 42.2& 40.9&{38.1}& 26.0& 30.2& 25.5& 22.3& 28.0& 26.7&23.6\\
        SD3~\cite{esser2024scaling}&38.9& 32.5&  26.3&20.7& 24.6& 36.6& 24.2& 14.3& 33.3& 24.3&16.3\\
        SVD~\cite{blattmann2023stable}&37.8& 36.6&  33.2&28.8& 24.9& 35.1& 26.2& 15.0& 33.7& 24.9&16.4\\
        V-JEPA2-G/16~\cite{assran2025v}& 40.1& 39.5& 36.3& 29.5& 25.4& 31.7& 26.7& 18.9& 31.4& 26.3&18.8\\ 
        \midrule
        HunyuanVideo~\cite{kong2024hunyuanvideo}    &45.3& 43.4& 40.9&36.4& 29.7& 44.6& 30.1& 16.6& 41.5& 28.7&19.3\\
 WAN-1.3B~\cite{wan2025wan}   &43.0& 41.3& 34.7&27.5& 25.8& 37.5& 27.2& 14.7& 36.8& 25.0&15.8\\
 WAN-14B~\cite{wan2025wan}   & 46.6& 45.1& 41.4&37.1& 32.2& 42.1& 34.5& 22.2& 41.7& {32.3}&{23.0}\\
 CogVideoX-2B~\cite{yang2024cogvideox}    & {48.2}& \underline{46.9}& \underline{42.8}&36.8& 32.0& \underline{47.1}& 33.2& 18.2& {43.9}& 31.3&21.2\\
 CogVideoX-5B~\cite{yang2024cogvideox}    & \textbf{49.7}& \textbf{48.6}& \textbf{45.2}&\underline{39.5}& \textbf{36.1}& \textbf{48.0}& \textbf{39.3}& \underline{23.3}& \textbf{47.5}& \textbf{36.0}&\textbf{25.0}\\
        \bottomrule
        \end{tabular}
    }
\end{table*}

\myparagraph{Zero-shot point tracking.}
To evaluate each backbone's inherent matching capability without task-specific training, we perform zero-shot tracking. We extract the query point feature from the first frame, compute matching scores against all spatial locations in subsequent frames, and take the soft argmax to form a trajectory. The trajectory is then upsampled to the original resolution via linear interpolation. We report $\delta_{\text{avg}}^x$, the average percentage of correctly tracked points across error distance thresholds of 1, 2, 4, 8, and 16 pixels. All models are evaluated at the same feature resolution for fair comparison, finding their layers that give the best tracking performance. For video DiTs, we follow the evaluation protocol of~\cite{nam2025emergent}, which computes matching costs from query-key projections within the 3D attention.

Further details about analysis settings are in Appendix~\ref{supp:analysis_setup}.

\begin{figure}[!t]
    \centering
    \includegraphics[width=\linewidth]{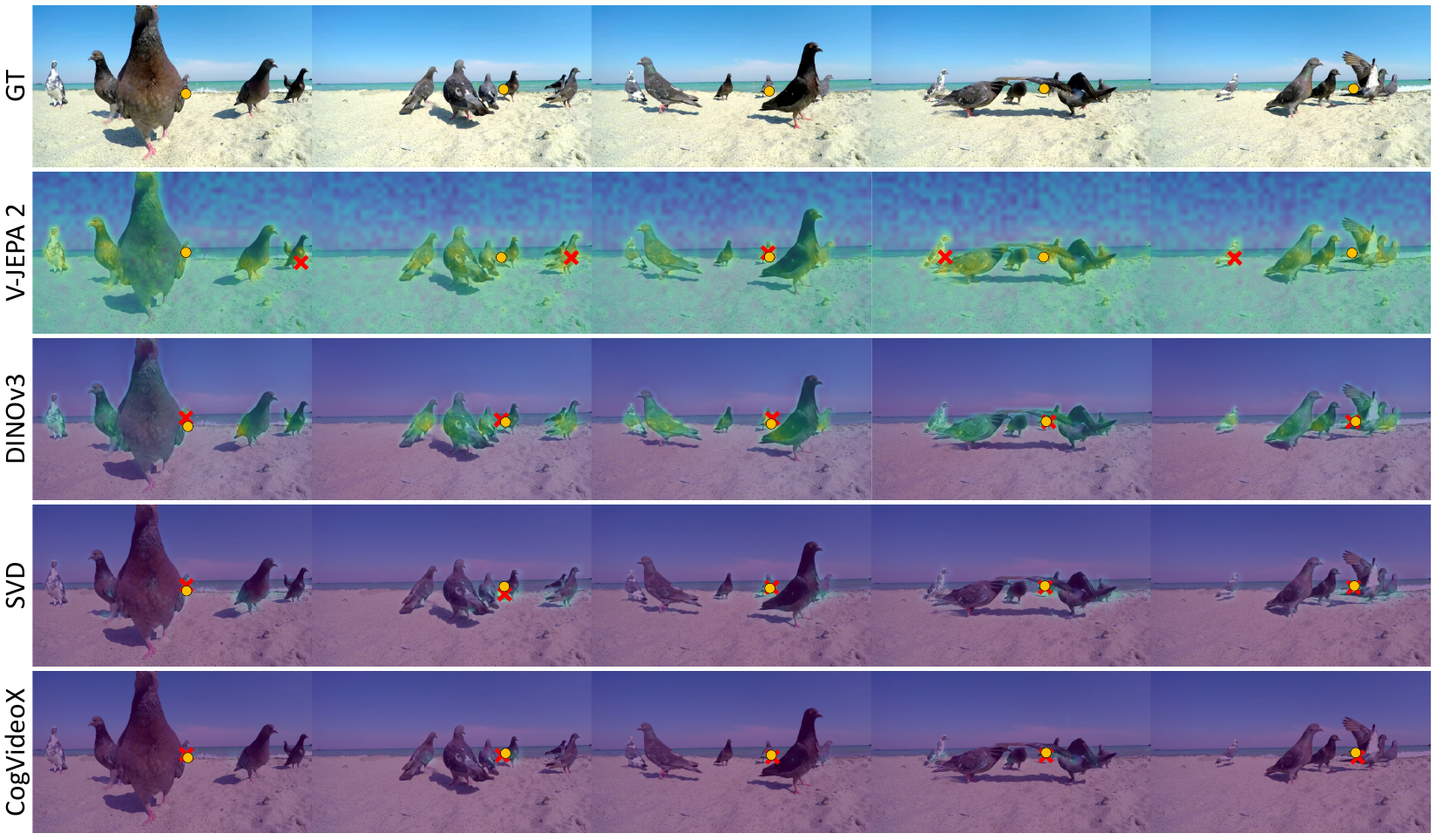}
   \caption{\textbf{Attention map visualization for ITTO-MOSE~\cite{demler2025tracker} on  VFMs.} Given a query point (\textcolor{orange}{orange dot} in the first frame), V-JEPA2~\cite{assran2025v}, DINOv3~\cite{simeoni2025dinov3}, and SVD~\cite{blattmann2023stable} produce noisy and scattered attention maps, failing to localize the corresponding point in subsequent frames (\textcolor{red}{red} $\times$). In contrast, CogVideoX-2B~\cite{yang2024cogvideox} produces sharper and more focused attention, enabling accurate point matching across frames.}
    \label{fig:zeroshot_analysis}
    \vspace{-10pt}
\end{figure}

\subsection{Analysis}
\label{subsec:findings}

\begin{finding}
\textbf{Finding 1.} Video DiTs provide the most robust feature representations for point tracking among VFMs.
\end{finding}

Video DiTs perform best among all VFMs and are comparable to supervised feature backbones, with CogVideoX-5B outperforming all comparisons. 
Although a perfectly controlled ablation is impossible give the variation across existing VFMs, CogVideoX-2B outperforms DINOv3-7B/16 despite its much smaller parameter count, suggesting that the key factor is temporal modeling capacity rather than model scale. \Cref{fig:zeroshot_analysis} further validates this by visualizing attention maps across VFMs. Given a query point in the first frame, other VFMs often fail to locate the corresponding point in later frames due to noisy and scattered attention, whereas CogVideoX-2B~\cite{yang2024cogvideox} produces sharper, more focused attention that enables accurate matching. We attribute this robustness to three factors of video DiTs, each supported by \Cref{tab:davis_zeroshot}.

\myparagraph{Large-scale real-world video data.} Large-scale real-world video pre-training exposes the model to diverse motion and appearance variations that image-based models such as the DINO family~\cite{oquab2023dinov2, simeoni2025dinov3} and SD3~\cite{esser2024scaling} never observe. Consistent with this, video DiTs surpass these image backbones. SD3 shares the DiT architecture but is trained only on images, and it lags well behind every video DiT, isolating pre-training data rather than architecture as the decisive factor. This mirrors~\cite{velez2025image}: video-trained diffusion consistently outperforms image-trained diffusion on point tracking.

\myparagraph{Full 3D attention.} It lets each token interact globally across space and time, whereas the decomposed attention of SVD~\cite{blattmann2023stable} cannot. Consequently, the full-attention DiTs (HunyuanVideo, CogVideoX, and WAN) consistently outperform SVD. This aligns with~\cite{yang2024cogvideox, lin2024open}, which show that full 3D attention enables global spatio-temporal interactions superior to decomposed 1+2D attention.

\myparagraph{Diffusion objective.} The MAE objective relies heavily on spatial cues while ignoring temporal relations, yielding sub-optimal matching representations~\cite{wu2023dropmae}. In contrast, diffusion models reconstruct all tokens across noise levels, producing richer correspondence representations~\cite{xiang2023denoising}. This suggests why CogVideoX-2B can outperform V-JEPA2.

\begin{finding}
\textbf{Finding 2.} Within the same model family, larger models consistently yield better tracking performance, with larger gains on challenging scenes.
\end{finding}

In \Cref{tab:davis_zeroshot}, across model families with multiple size variants, \eg DINOv3 (B/16 and 7B/16), WAN (1.3B and 14B), and CogVideoX (2B and 5B), we find that larger variants consistently achieve higher tracking performance.

The scaling effect is particularly evident under challenging conditions. Under motion blur on DAVIS, larger models exhibit greater robustness: WAN-14B degrades by 8.0\%p (45.1\% $\rightarrow$ 37.1\%) from severity 1 to 5, whereas WAN-1.3B drops by 13.8\%p (41.3\% $\rightarrow$ 27.5\%). This suggests that larger models learn more robust representations for challenging tracking scenarios.
\section{DiTracker: Leveraging Video DiTs for Point Tracking}
\label{sec:method}
\begin{figure*}[t]
    \centering
    \includegraphics[width=\textwidth]{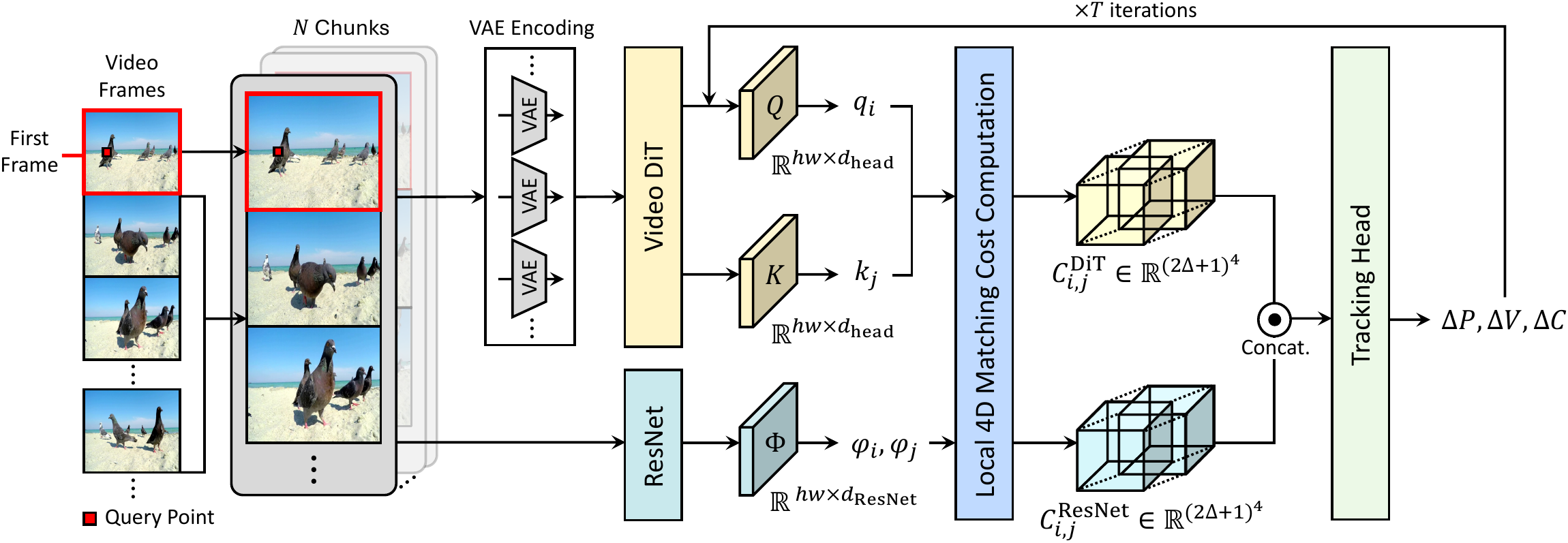}
   \caption{\textbf{Overall Architecture of DiTracker.} For long video sequences, input frames are divided into $N$ chunks with the global first frame prepended. Individual video frames are then encoded via VAE. These encoded latents are processed by a video DiT to extract query feature $q_i$ and key feature $k_j$, which are then used to compute a hierarchical local 4D matching cost $\mathcal{C}^{\text{DiT}}_{i,j}$. The video DiT local cost is subsequently fused with the ResNet local cost $\mathcal{C}^{\text{ResNet}}_{i,j}$. Finally, a tracking head refines the trajectories over $T$ iterations, updating displacement ($\Delta P$), visibility ($\Delta V$), and confidence ($\Delta C$). }
    \label{fig:arch}
\end{figure*}

Based on the findings in \Cref{sec:analysis}, we introduce \textbf{DiTracker}, a framework that integrates video DiT features into existing tracking paradigms~\cite{karaev2024cotracker3, cho2024local, harley2025alltracker}, as illustrated in \Cref{fig:arch}. Rather than improving the tracking head, we focus on the three components that turn a generative video DiT into an effective tracking backbone: query-key matching cost computation, ResNet cost fusion, and LoRA adaptation. The remaining architectural details follow standard tracking pipelines and are provided in Appendix~\ref{supp:arch_details}.

\subsection{Preliminaries} 
\label{sec:prelim}
\myparagraph{Point Tracking.} Given a video sequence $\mathbf{X} \in \mathbb{R}^{F \times H \times W \times 3}$ with $F$ frames at spatial resolution $H \times W$ and a query point $\mathbf{p} = (x, y)$ in a query frame, point tracking aims to predict the corresponding positions $\{P_i = (x_i, y_i)\}_{i=1}^F$ across all frames, together with visibility $\{V_i \in [0,1]\}_{i=1}^F$ and confidence $\{C_i \in [0,1]\}_{i=1}^F$.

\myparagraph{Video Diffusion Transformers (DiTs).} Video DiTs consist of a VAE and a denoising transformer. Given a video sequence $X \in \mathbb{R}^{F \times H \times W \times 3}$, the VAE encodes $X$ into latent representations
$\mathbf{z}_{\text{video}} \in \mathbb{R}^{f \times h \times w \times d_{\text{video}}}$,
where $(F,H,W,3)$ denote the number of frames, height, width, and RGB channels, and
$(f,h,w,d_{\text{video}})$ denote the corresponding latent dimensions.
During the diffusion process, Gaussian noise $\boldsymbol{\epsilon} \sim \mathcal{N}(0, I)$ is added to $\mathbf{z}_{\text{video}}$,
and the denoising transformer $v_\theta$ learns to predict the velocity field that maps noisy latents back to clean ones via flow matching~\cite{lipman2022flow}.

\subsection{Video DiT feature extraction}
\label{subsec:feature_extraction}

Following~\cite{nam2025emergent}, we encode each frame $X_i$ using a pre-trained VAE encoder to obtain frame latent $z_i$. We then extract query and key features $Q_i^{l,m}, K_i^{l,m} \in \mathbb{R}^{H' \times W' \times d}$ from the full 3D attention of video DiT at layer $l$ and head $m$. Here, $H' \times W'$ denotes the spatial resolution in the DiT latent space and $d$ is the head dimension. The layer and head are selected based on zero-shot tracking performance. To efficiently process long videos, we divide the video into $N$ chunks and prepend the first frame $X_1$ as a shared anchor to each chunk, ensuring consistent query-key distributions.

\subsection{Query-key matching cost}
\label{subsec:cost}
The first component, query-key matching cost computation, derives the matching cost directly from the query and key projections of the video DiT. Unlike hidden states, these projections already encode inter-frame correspondences within the full 3D attention, so we reuse the model's internal matching mechanism instead of learning a new one. Given the local query feature $q_i$ sampled within a radius $\Delta$ around the query point in frame $i$, and the local key feature $k_j$ sampled within the same radius $\Delta$ around the current estimate in target frame $j$, we form the matching cost with a scaled dot-product attention:
\begin{equation}
\label{equation:local}
\mathcal{C}_{i,j}^{\text{DiT}} = \text{Softmax}\!\left(\frac{q_i\, k_j^\top}{\sqrt{d_\text{head}}}\right).
\end{equation}
To capture relationships at multiple scales, we build this cost over a local feature pyramid, following standard tracking pipelines~\cite{karaev2024cotracker3, harley2025alltracker}.

\subsection{Matching cost fusion}
\label{subsec:fusion}
Since video DiTs operate in a spatially compressed latent space (\eg, $H/16 \times W/16$), they often lack the fine-grained spatial detail required by standard tracking heads (\eg, $H/4 \times W/4$ for CoTracker3~\cite{karaev2024cotracker3} and $H/8 \times W/8$ for AllTracker~\cite{harley2025alltracker}). A naive solution is to train an upsampler on the query and key features, but we find that this disrupts the pretrained matching distribution induced by the full 3D attention. To resolve this, we instead fuses the DiT matching cost with an additional ResNet branch that provides higher-resolution features. We extract ResNet features $\Phi_i = \text{ResNet}(X_i)$ and compute an analogous ResNet matching cost $\mathcal{C}_{i,j}^{\text{ResNet}}$.
Both costs are then flatten, concatenated, and projected into a cost embedding $E_j$ via an MLP:
\begin{equation}
\label{equation:fusing}
E_j = \text{MLP}\!\left(\left[\text{Flatten}(\mathcal{C}_{i,j}^{\text{DiT}}),\; \text{Flatten}(\mathcal{C}_{i,j}^{\text{ResNet}})\right]\right) \in \mathbb{R}^{d_E}.
\end{equation}
Cost-level fusion preserves the pretrained matching distribution of the DiT, while the ResNet branch recovers details that the lower-resolution DiT features cannot provide.

\subsection{Trajectory estimation}
\label{subsec:trajectory}
Given the cost embeddings, we estimate point trajectories with the iterative transformer-based refinement module of CoTracker3~\cite{karaev2024cotracker3}. Over $T$ iterations, it repeatedly predicts residual updates to position ($\Delta P$), visibility ($\Delta V$), and confidence ($\Delta C$), and resamples cost features at the refined locations.

\subsection{Training}
\label{subsec:training}

The third component adapts the video DiT backbone for tracking with LoRA~\cite{hu2022lora}. LoRA inserts trainable low-rank matrices into the attention layers while keeping the backbone frozen, enabling efficient transfer of generative features to tracking-aligned representations. The model is trained with the combined objective:
\begin{equation}
    \mathcal{L} = \mathcal{L}_{\text{track}} + \mathcal{L}_{\text{vis}} + \mathcal{L}_{\text{conf}},
    \label{eq:loss}
\end{equation}
where $\mathcal{L}_{\text{track}}$, $\mathcal{L}_{\text{vis}}$, and $\mathcal{L}_{\text{conf}}$ are the trajectory regression, visibility classification, and confidence losses, respectively, summed across iterative refinement steps.

\begin{table*}[t]
\caption{\textbf{Quantitative evaluation on the TAP-Vid and ITTO-MOSE benchmarks.} \textbf{Bold} and \underline{underline} denote the best and second-best results. The bottom row reports the performance gain of DiTracker (CoTracker3) over CoTracker3. $\dagger$: results reported in the original paper.}
\centering
\resizebox{\textwidth}{!}{
\begin{tabular}{l|c|c|c|c|ccc|ccc|ccc|ccc|ccc}
\toprule
\multirow{2}{*}{Method} &  \multirow{2}{*}{\shortstack{Feature\\Backbone}}&\multirow{2}{*}{\shortstack{Training\\Dataset}} & \multirow{2}{*}{Steps} &\multirow{2}{*}{Batch}& \multicolumn{3}{c|}{ITTO-MOSE} & \multicolumn{3}{c|}{DAVIS} & \multicolumn{3}{c|}{Kinetics} & \multicolumn{3}{c|}{RoboTAP}& \multicolumn{3}{c}{RGB-Stacking}\\
&  && & & AJ$\uparrow$ & $<\delta^x_\text{avg}\uparrow$ & OA$\uparrow$   & AJ$\uparrow$ & $<\delta^x_\text{avg}\uparrow$ & OA$\uparrow$  & AJ$\uparrow$ & $<\delta^x_\text{avg}\uparrow$ &OA$\uparrow$  & AJ$\uparrow$ & $<\delta^x_\text{avg}\uparrow$ & OA$\uparrow$  & AJ$\uparrow$ & $<\delta^x_\text{avg}\uparrow$ &OA$\uparrow$  \\
\midrule \midrule
TAPIR~\cite{doersch2023tapir} &  ResNet&Kub.& 50k& 4& 33.1& 46.1& {75.9}& 56.2& 70.0& 86.5& 49.6& 64.2&85.0 & 59.6& 73.4& 87.0& 55.5& 69.7&88.0\\
BootsTAPIR~\cite{doersch2024bootstap} &  ResNet&Kub.+Real(\textbf{15M})& 200k& 1,536& 36.9& 51.1& 76.4 & 61.4& 73.6& {88.7}& {54.6}& {68.4}&{86.5} & 64.9& 80.1& 86.3& 70.8& 83.0&89.9\\
TAPTRv2~\cite{li2024taptrv2}& ResNet &Kub.& 44k& 32& 36.3& 47.8& 76.9& {63.0}& 76.1& {91.1}& 49.0& 64.4&{85.2} & 60.9& 74.6& 87.7& 52.9& 71.2 & 79.0\\
LocoTrack~\cite{cho2024local} &  ResNet&Kub.& 400k&8& {39.2}& {52.3}& {78.2}& 64.8& {77.4}& 86.2& 52.3& 66.4&82.1 & 62.3& 76.2& 87.1& 69.7& 83.2&89.5\\
 SptialTrackerv2~\cite{xiao2025spatialtrackerv2}& ResNet &Mix& 300k+ & - & 23.7& 46.6& 51.2& {63.3}& 74.0& 89.7& -& -&- & 64.0& 76.9& 89.1& 67.9& 81.4&88.9\\
TAPNext~\cite{zholus2025tapnext}&  TRecViT&Kub.& 300k& 256 & 35.0 & 50.2 & 77.1 & 62.4& 76.6& 90.5& 53.3& 67.9&87.0 & 59.5& 72.8& 88.0& 76.4& 86.3&\textbf{95.5}\\
 AllTracker~\cite{harley2025alltracker}& ConvNeXt& {Kub.+Mix(27k)}&600k& 8& 44.4& \underline{59.2}& 80.5& 63.3& 76.3& 90.1& 53.9& 66.5&89.4& 71.9& \textbf{83.4}& 92.8& \underline{81.1}& 90.0&92.8\\
 Track-On2~\cite{aydemir2025track}&  DINOv3&Kub.& 36k& 32& 42.2& \textbf{59.3}& \textbf{81.2}& \textbf{67.0}& \textbf{79.9}& \underline{92.0}& 55.3& 69.3& \underline{89.6}& 68.1& 80.5& \underline{93.4}& 62.7& 82.2&79.8\\
 CoWTracker$^\dagger$~\cite{lai2026cowtracker}&  VGGT&Kub.& 50k& 32& -& -& -& \underline{65.5}& 78.0& \textbf{92.1}& \textbf{60.9}& \textbf{73.1}& \textbf{91.5}& \textbf{73.2}& \textbf{83.4}& \textbf{94.7}& \textbf{85.4}& \textbf{92.8}&73.2\\

 \midrule
 CoTracker3-Kub~\cite{karaev2024cotracker3}&  ResNet&Kub.& 50k& 32 & 34.0& 47.1& 71.7& 57.8& 74.9& 80.4& 49.3& 62.7&78.7 & 59.9& 73.4& 71.6& 74.0& 84.9&90.5\\
 CoTracker3~\cite{karaev2024cotracker3}&  ResNet&{Kub.+Real(15k)}& 65k& 32 & 42.4& {55.8}& 80.4& 64.8& {76.8}& 91.7& {54.7}& {67.8}&87.5& 64.7& 78.0& 89.4& 74.3& 85.2&\underline{92.4}\\
 \rowcolor{gray!15} \textbf{DiTracker-Kub(CoTracker3)}&  CogVideoX-2B&{Kub.} & 39k& 4  & 43.7& 58.0& 79.3& 62.9& 77.3& 85.6& 53.2& 65.9&83.4 & 65.0& 76.2& 88.0& 72.2& 82.0&88.5\\
 \rowcolor{gray!15} \textbf{DiTracker(CoTracker3)}&  CogVideoX-2B&{Kub.}+PO+DR(0.6k)& 39k&4  & \textbf{45.1}& 59.0& \underline{80.6}    & 64.8& {78.0}&89.0& 55.2& {68.3}& 86.3 & 66.5& 78.7& 90.0& 74.8& 85.2&90.9\\
 \rowcolor{gray!15} $\Delta$ vs.\ CoTracker3 & & & & & \up{2.7}& \up{3.2}& \up{0.2}& \eqd{+0.0}& \up{1.2}& \dn{2.7}& \up{0.5}& \up{0.5}& \dn{1.2}& \up{1.8}& \up{0.7}& \up{0.6}& \up{0.5}& \eqd{+0.0}& \dn{1.5}\\
\bottomrule
\end{tabular}
}
\label{tab:main_quan}
\end{table*}

\begin{table*}[t]
\caption{\textbf{Quantitative evaluation on TAP-Vid-DAVIS~\cite{doersch2022tap} with common corruptions from ImageNet-C~\cite{hendrycks2019benchmarking}.} Each entry reports $\delta^x_{\text{avg}}$ at corruption severity 2. \textbf{Bold} and \underline{underline} denote the best and second-best results. The bottom row reports the performance gain of DiTracker (CoTracker3) over CoTracker3.}
\centering
\label{tab:corrupted}
\resizebox{0.9\textwidth}{!}{
\begin{tabular}{l|ccc|cccc|cccc|c}
\toprule 
 \multirow{2}{*}{Method} & \multicolumn{3}{c|}{Noise} & \multicolumn{4}{c|}{Blur} & \multicolumn{4}{c|}{Weather}  & \multirow{2}{*}{Avg.} \\
   & Gauss. & Shot & Impulse & Defocus & Glass & Motion & Zoom & Snow & Frost & Fog & Bright &  \\
   \midrule \midrule
   TAPIR \cite{doersch2023tapir} & 59.6& 59.1& 58.5& 62.7& 61.7& 57.0& 53.5& 61.8& 58.4& 63.8& 69.8&60.5
\\
 BootsTAPIR \cite{doersch2024bootstap} & 53.1& {67.1}& 64.5& 64.7& 63.5& 59.9& 56.7& 66.9& 67.5& 71.0& 73.8&64.4
\\
 TAPTRv2 \cite{li2024taptr}& 64.1& 61.8& 61.7& 64.7& 60.7& 56.0& 55.7& 62.5& 67.1& 72.4& 75.3&63.8
\\
 LocoTrack \cite{cho2024local} & {66.7}& 66.4& {65.8}& {68.4}& {66.5}& {63.6}& {58.4}& {68.0}& {70.1}& {74.1}& {75.1}&67.6
\\
 SptialTrackerv2~\cite{xiao2025spatialtrackerv2}& 61.6& 62.3& 61.0& 65.6& 63.2& 58.1& 56.7& 63.6& 64.7& 69.0& 73.6&63.6
\\
 TAPNext~\cite{zholus2025tapnext}& \textbf{75.4}& \textbf{75.9}& \textbf{74.7}& 70.2& \underline{75.0}& \textbf{67.4}& 58.9& 63.9& 63.6& 70.8& 77.1&70.2\\
 AllTracker~\cite{harley2025alltracker}& 64.5& 63.4& 65.9& \textbf{75.1}& \textbf{75.7}& \underline{66.4}& \textbf{60.7}& 72.0& \underline{74.5}& 76.5& 77.1&\underline{70.7}\\
 Track-On2~\cite{aydemir2025track} & 69.0& 66.7& 70.6& 69.8& 60.7& 53.7& 58.9& \textbf{78.8}& \textbf{79.3}& \textbf{78.8}& \textbf{79.4}&69.7\\

 \midrule
 CoTracker3-Kub~\cite{karaev2024cotracker3}& 54.4& 51.7& 51.8& 66.1& 60.2& 57.3& 55.5& 62.6& 64.7& 73.7& 73.6&61.1
\\
 CoTracker3~\cite{karaev2024cotracker3}& {68.5}& {68.0}& {67.6}& \underline{70.7}& {68.5}& {64.3}& \underline{60.2}& {70.8}& {71.6}& 75.8& {75.3}& 69.2
\\
 \rowcolor{gray!15} \textbf{DiTracker-Kub(CoTracker3)}& 70.5& 70.7& 69.8& \underline{71.0}& 64.8& 63.7& {60.2}& 70.4& {72.5}& 75.5& 76.3&69.6\\
 \rowcolor{gray!15} \textbf{DiTracker(CoTracker3)}& \underline{72.8}& \underline{72.1}& \underline{71.7}& \underline{71.0}& {67.1}& {65.0}& \underline{60.4}& \underline{73.1}& 74.2& \underline{76.8}& \underline{77.5}&\textbf{71.1}\\
 \rowcolor{gray!15} $\Delta$ vs.\ CoTracker3 & \up{4.3}& \up{4.1}& \up{4.1}& \up{0.3}& \dn{1.4}& \up{0.7}& \up{0.2}& \up{2.3}& \up{2.6}& \up{1.0}& \up{2.2}& \up{1.9}\\
 \bottomrule
\end{tabular}} 
\end{table*}

\section{Experiments}

\subsection{Implementation details} We train DiTracker with CogVideoX-2B~\cite{yang2024cogvideox} as the video DiT backbone and CoTracker3~\cite{karaev2024cotracker3} as the tracking head. DiTracker-Kub is trained exclusively on Kubric~\cite{greff2022kubric}, while DiTracker is fine-tuned on Kubric augmented with 655 synthetic samples from PointOdyssey~\cite{zheng2023pointodyssey} and DynamicReplica~\cite{karaev2023dynamicstereo}. Additional details are in Appendix~\ref{supp:method_imde}.

\subsection{Evaluation settings}
We evaluate our model on TAP-Vid-DAVIS~\cite{doersch2022tap}, TAP-Vid-Kinetics~\cite{doersch2022tap}, ITTO-MOSE~\cite{demler2025tracker}, RoboTAP~\cite{vecerik2024robotap}, and TAP-Vid-RGB-Stacking~\cite{doersch2022tap}, with per-benchmark details in Appendix~\ref{supp:eval_protocol}.
For robustness evaluation, we apply ImageNet-C~\cite{hendrycks2019benchmarking} corruptions to TAP-Vid-DAVIS, including noise, blur, and weather. 
Following standard practice, all videos are resized to $256 \times 256$ and interpolated to each model's input resolution for feature extraction.

We adopt the TAP-Vid metrics~\cite{doersch2022tap}: $\delta_{\text{avg}}^{x}$, the percentage of correctly tracked visible points averaged over distance thresholds of 1, 2, 4, 8, and 16 pixels; Occlusion Accuracy (OA), the binary classification accuracy of the visible-versus-occluded prediction; and Average Jaccard (AJ), which jointly scores trajectory accuracy and visibility prediction.

\begin{table*}[t]
\caption{\textbf{Ablation studies on adapting video DiT features for tracking.} We separately ablate (a) the matching descriptor, (b) cost fusion with the ResNet branch, and (c) LoRA adaptation, all evaluated on TAP-Vid-DAVIS~\cite{doersch2022tap}. \textbf{Bold} marks the best result in each study, and the configuration adopted by DiTracker is highlighted in gray.}
\centering
\footnotesize
\setlength{\tabcolsep}{4pt}
\begin{subtable}[t]{0.27\textwidth}
\centering
\begin{tabular}{l|ccc}
\toprule
Descriptor & AJ & $<\delta^x_\text{avg}$ & OA \\
\midrule \midrule
Hidden state & 45.7 & 63.5 & 70.9 \\
\rowcolor{gray!15} Query-Key & \textbf{52.6} & \textbf{67.5} & \textbf{78.1} \\
\bottomrule
\end{tabular}
\caption{\textbf{Feature selection.} Query-key projections outperform hidden states as the matching descriptor.}
\label{tab:qk_ablation}
\end{subtable}
\hfill
\newsavebox{\costfusionbox}%
\savebox{\costfusionbox}{%
\begin{tabular}{ll|ccc}
\toprule
& Method & AJ & $<\delta^x_\text{avg}$ & OA \\
\midrule \midrule
(a) & Only DiT & 52.7 & 68.5 & 78.3 \\
(b) & Upsampler & 43.0 & 65.5 & 66.5 \\
\midrule
(c) & Feat. Concat & \textbf{56.4} & 71.1 & \textbf{81.0} \\
(d) & Cost Sum & 55.4 & 70.0 & 79.9 \\
\rowcolor{gray!15} (e) & Cost Concat & 56.0 & \textbf{72.2} & 79.2 \\
\bottomrule
\end{tabular}}%
\begin{subtable}[t]{\wd\costfusionbox}
\centering
\usebox{\costfusionbox}
\caption{\textbf{Cost fusion.} Concatenating DiT and ResNet matching costs performs best.}
\label{tab:resnet_ablation}
\end{subtable}
\hfill
\begin{subtable}[t]{0.31\textwidth}
\centering
\begin{tabular}{ll|c|ccc}
\toprule
& Layer & Rank & AJ & $<\delta^x_\text{avg}$ & OA \\
\midrule \midrule
(a) & None & - & 41.0 & 57.5 & 65.3 \\
(b) & 17 & 128 & 42.9 & 59.9 & 66.5 \\
(c) & [0, 18) & 2 & 44.4 & 65.7 & 72.7 \\
(d) & [0, 18) & 32 & 49.8 & \textbf{67.9} & 72.7 \\
\rowcolor{gray!15} (e) & [0, 18) & 128 & \textbf{52.6} & 67.5 & \textbf{78.1} \\
\bottomrule
\end{tabular}
\caption{\textbf{LoRA adaptation.} Broader layer coverage and higher rank improve adaptation; all-layer rank-128 is best.}
\label{tab:abl_lora}
\end{subtable}
\label{tab:ablation}
\end{table*}

\begin{figure*}[t!]
    \centering
    \includegraphics[width=\linewidth]{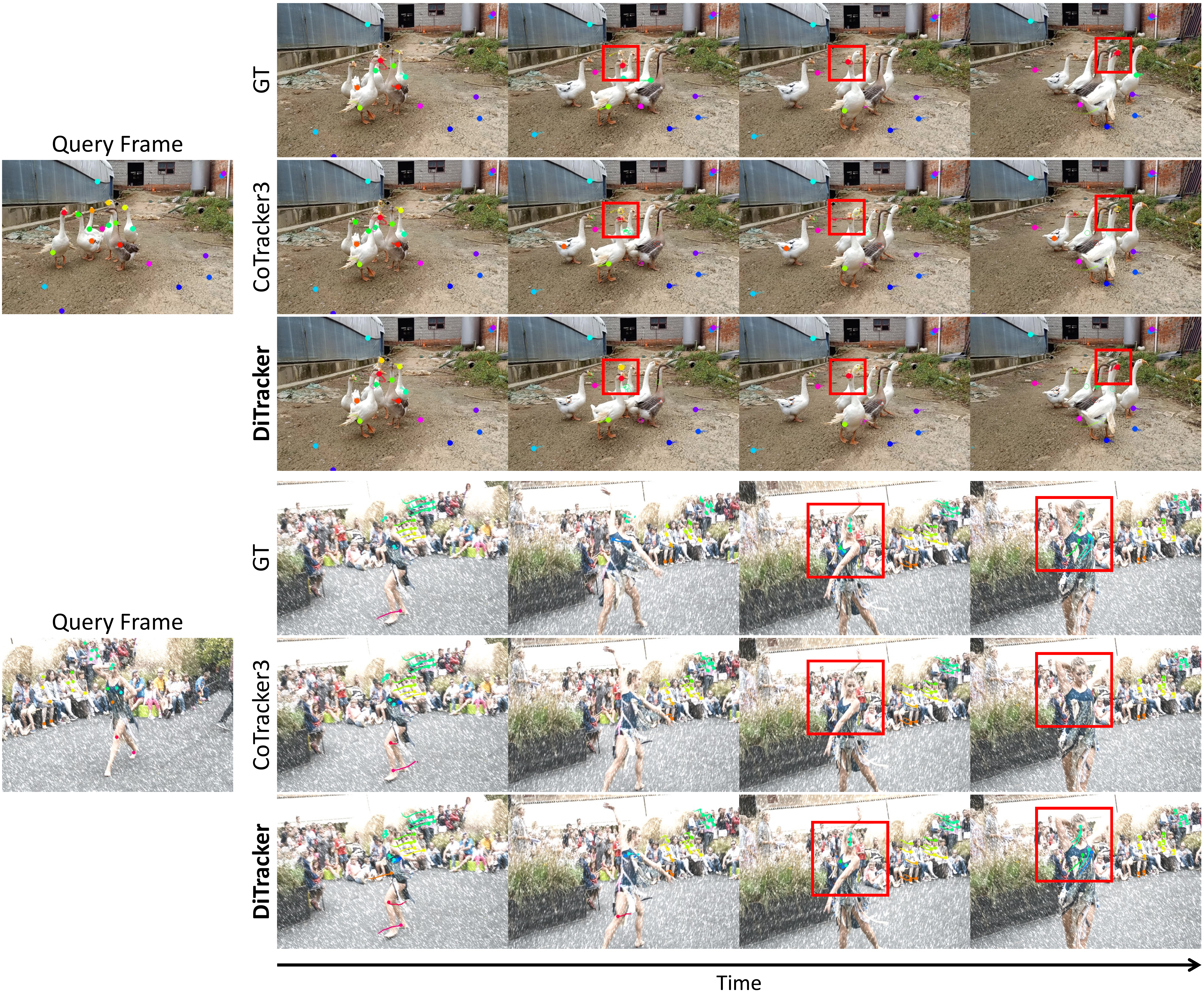}
    \caption{\textbf{Qualitative results on the ITTO-MOSE~\cite{demler2025tracker} and corrupted TAP-Vid-DAVIS benchmarks.} The TAP-Vid-DAVIS sample is corrupted with the snow perturbation. Compared to CoTracker3~\cite{karaev2024cotracker3}, our DiTracker predicts more accurate point trajectories under challenging real-world scenarios such as dynamic motions and occlusions.}
    \label{fig:main_qual}
\end{figure*}

\subsection{Results}

\myparagraph{Quantitative result.}
Our most direct competitor is CoTracker3, which shares the same tracking head as DiTracker and differs only in the feature backbone, isolating the backbone's effect. On the standard benchmarks in \Cref{tab:main_quan}, DiTracker is competitive with recent state-of-the-art methods and consistently outperforms CoTracker3. Under Kubric-only training, DiTracker-Kub surpasses CoTracker3-Kub by 10.9\%p $\delta_\text{avg}^x$ on ITTO-MOSE and 2.4\%p on TAP-Vid-DAVIS. More notably, DiTracker fine-tuned with only 655 synthetic samples matches or surpasses CoTracker3 across all five benchmarks, even though CoTracker3 uses 15k additional real-world videos, improving $\delta_\text{avg}^x$ by 2.2\%p on ITTO-MOSE and reaching 45.1\% AJ. Since the backbone is the only difference, a video DiT backbone can directly replace real-video supervision under the same head.

This advantage holds under visual degradation. In \Cref{tab:corrupted}, DiTracker attains the best average score (71.1\%) on corrupted TAP-Vid-DAVIS, surpassing CoTracker3 (69.2\%) despite its real-world video training, consistent with the degradation-invariant behavior in \Cref{subsec:findings}.

\myparagraph{Qualitative result.} \Cref{fig:main_qual} visualizes point trajectories on ITTO-MOSE and corrupted TAP-Vid-DAVIS, where DiTracker tracks robustly through large motions and recovers point identity after occlusion. Additional visualizations are in Appendix~\ref{supp:visualize}.

\begin{table}[t]
\centering
\caption{\textbf{Generalization across tracking heads and scalability with backbone size.} ``Corr.\ DAVIS'' reports $\delta^x_\text{avg}$ on TAP-Vid-DAVIS averaged over all ImageNet-C corruptions at severity 2. \textbf{Bold} and \underline{underline} denote the best and second-best results within each block.}
\label{tab:abl_general}
\resizebox{\linewidth}{!}{
\begin{tabular}{l|ccc|ccc|c}
\toprule
\multirow{2}{*}{Method} & \multicolumn{3}{c|}{ITTO-MOSE} & \multicolumn{3}{c|}{TAP-Vid-DAVIS} & Corr. DAVIS \\
 & AJ & $<\delta^x_\text{avg}$ & OA & AJ & $<\delta^x_\text{avg}$ & OA & Avg. \\
\midrule \midrule
\multicolumn{8}{l}{\textit{Backbone scale (CoTracker3 head)}} \\
DiTracker(WAN-1.3B) & 37.4 & 52.9 & 74.6 & 52.7 & 70.1 & 74.7 & 62.9 \\
DiTracker(WAN-14B) & \textbf{39.9} & \textbf{54.0} & \textbf{77.0} & \textbf{57.0} & \textbf{72.0} & \textbf{79.4} & \textbf{64.7} \\
\midrule
\multicolumn{8}{l}{\textit{Tracking head (WAN-1.3B backbone)}} \\
AllTracker-Kub~\cite{harley2025alltracker} & 43.7 & 57.7 & \textbf{80.5} & 61.8 & 75.3 & 87.8 & 65.7 \\
AllTracker~\cite{harley2025alltracker} & 44.4 & \textbf{59.2} & \textbf{80.5} & 63.3 & 76.3 & \textbf{90.1} & \underline{70.7}\\
DiTracker(AllTracker) & \textbf{44.6} & \underline{58.8}& \underline{80.2}& \textbf{64.4} & \textbf{77.9} & \underline{89.1}& \textbf{71.4} \\
\bottomrule
\end{tabular}}
\end{table}

\subsection{Architecture ablation}
To identify the effective design choices for leveraging a video DiT as a feature backbone, we conduct ablation experiments with batch size 1, trained for 5k steps.

\myparagraph{Video DiT feature selection.}
We examine how to effectively leverage video DiT features for supervised tracking. Table~\ref{tab:qk_ablation} compares feature descriptors: query-key projections achieve 52.6\% AJ, outperforming hidden state features (45.7\% AJ) by 6.9\%p. Query-key projections directly encode inter-frame correspondences within the attention mechanism, while hidden states dilute this matching prior through post-attention aggregation.

\myparagraph{Matching cost fusion.}
Table~\ref{tab:resnet_ablation} compares strategies for combining DiT and ResNet matching costs. Using only DiT features with bilinear interpolation {(a)} serves as the baseline. Attaching a learnable upsampler to the query-key features {(b)} degrades AJ from 52.7\% to 43.0\%, as the upsampler creates a gradient shortcut that bypasses LoRA adaptation. Among ResNet fusion strategies (c, d, e), which all improve over the baseline, cost concatenation {(e)} achieves the highest $\delta_\text{avg}^x$ at 72.2\%, allowing the tracking head to learn adaptive weighting between cost sources.

\myparagraph{LoRA adaptation.}
In Table~\ref{tab:abl_lora}, we observe that LoRA adaptation is essential for transferring generative DiT features to tracking. Without LoRA, frozen DiT features yield only 41.0\% AJ (a). Applying LoRA improves AJ by 1.9\%p. (b) and (c) further show that extending LoRA across all attention layers substantially outperforms single-layer adaptation. (c) and (d) show that increasing the LoRA rank from 2 to 128 yields consistent gains.

\subsection{Generalization and scalability}
\myparagraph{Scaling the backbone.} We verify that Finding 2 in \Cref{subsec:findings}, that larger video DiTs within a family yield better zero-shot tracking, persists after training by pairing a CoTracker3~\cite{karaev2024cotracker3} head with WAN2.1-1.3B and WAN2.1-14B~\cite{wan2025wan} and matching the number of learnable parameters (LoRA rank 64 for 1.3B and 16 for 14B). As shown in the top block of Table~\ref{tab:abl_general}, WAN2.1-14B outperforms WAN2.1-1.3B by 4.3\%p AJ on ITTO-MOSE and 2.5\%p AJ on TAP-Vid-DAVIS, and is also more robust under corruptions. These experiments use batch size 1 and 11k training steps.

\myparagraph{Switching the tracking head.} The benefit of video DiT features should not be tied to a single head, so we replace the ConvNeXt feature backbone of the recent AllTracker~\cite{harley2025alltracker} with our WAN-1.3B~\cite{wan2025wan} backbone and train only on Kubric~\cite{greff2022kubric} with batch size 4 and 128k steps, which we denote DiTracker(AllTracker). As shown in the bottom block of Table~\ref{tab:abl_general}, it improves over AllTracker-Kub across all benchmarks, raising AJ by 2.6\%p on TAP-Vid-DAVIS and the corrupted-DAVIS average by 5.7\%p. Despite using only Kubric and far fewer training steps, it also matches or surpasses AllTracker trained on 27k mixed data with 600k steps on most metrics, including ITTO-MOSE AJ and the corrupted-DAVIS average. Since this improvement holds for a head with a distinct architecture, the benefit of video DiT features is not tied to a single tracking head, confirming that they form a robust, scale-friendly, and head-agnostic backbone for point tracking.

\section{Conclusion}

We study video DiTs as robust feature backbones for point tracking. Our systematic analysis of visual foundation models shows that pre-trained video DiT features are more robust to real-world challenges than other models and even surpass ResNet backbones supervised on tracking data.
We suggest that this robustness stems from large-scale video pretraining, full 3D attention, and the diffusion objective. Building on this, we propose \textbf{DiTracker}, which adapts these features through query-key matching cost, cost-level fusion with a ResNet branch, and LoRA. Under the same tracking head, DiTracker is trained only on synthetic data with far fewer iterations, yet outperforms CoTracker3 trained on real-world videos, with the largest gains on challenging and corrupted scenes. It also generalizes across tracking heads and scales with backbone size, confirming video DiTs as a robust, data-efficient, and scalable backbone for point tracking. More broadly, our results suggest that generative video pretraining provides strong real-world priors for correspondence, letting future advances in video generation directly benefit point tracking.

\clearpage

{
    \small
    \bibliographystyle{ieeenat_fullname}
    \bibliography{main}
}

\clearpage
\section*{\large Supplementary material}


\vspace{6pt}
\myparagraph{Overview.}
This supplementary material provides additional details and experimental results. \Cref{supp:analysis_setup} describes the VFMs, point tracking models and benchmarks used in our zero-shot analysis, along with the feature extraction and trajectory estimation procedures. \Cref{supp:arch_details} details the full architecture of DiTracker with training loss functions. \Cref{supp:method_imde} lists implementation details and hyperparameters. \Cref{supp:visualize} presents additional qualitative results. Finally, \Cref{supp:limitation} discusses future work.

\section{Analysis Setup}
\label{supp:analysis_setup}

\subsection{Visual Foundation Models (VFMs) Details}
\label{supp:vfm_details}
\myparagraph{DINO.}
DINOv2~\cite{oquab2023dinov2} is a self-supervised vision transformer trained on a curated dataset of 142M images using a combination of self-distillation and masked image modeling objectives. It produces versatile dense visual features without any task-specific supervision. DINOv3~\cite{simeoni2025dinov3} extends DINOv2 by scaling both the training dataset and model size, and introduces Gram-based feature anchoring to improve dense feature quality. We use the ViT-B variant for both models, and additional ViT-7B variant for DINOv3.

\myparagraph{SD3.} Stable Diffusion 3 (SD3)~\cite{esser2024scaling} is a text-to-image diffusion model based on the Multimodal Diffusion Transformer (MMDiT) architecture. It processes image and text tokens through separate streams with joint attention, trained using a rectified flow formulation. We include SD3 as a representative image diffusion model to examine whether image-level generative priors transfer to temporal correspondence.

\myparagraph{V-JEPA2.}
V-JEPA2~\cite{assran2025v} is a self-supervised video model that learns representations through masked feature prediction in latent space, extending V-JEPA by scaling to 1M hours of video data and up to 8B parameters. It achieves strong performance across video understanding tasks without relying on pixel-level reconstruction or text supervision. We use the ViT-G variant in our evaluation.

\myparagraph{SVD.}
Stable Video Diffusion (SVD)~\cite{blattmann2023stable} is a UNet-based latent video diffusion model trained through a three-stage pipeline: text-to-image pretraining, video pretraining on a large curated dataset, and high-quality video finetuning. It employs temporal convolution and attention layers incorporated into the image UNet architecture.

\myparagraph{HunyuanVideo.}
HunyuanVideo~\cite{li2024hunyuan} is an open-source video generation model with 13B parameters, built on a diffusion transformer architecture with full 3D attention. It employs a dual-stream to single-stream design for text-video interaction and is trained through progressive fine-tuning from image to video generation.

\myparagraph{CogVideoX.}
CogVideoX~\cite{yang2024cogvideox} is a DiT-based text-to-video generation model that uses a 3D variational autoencoder for spatiotemporal compression and an expert transformer with full 3D attention over video and text tokens. It is available in 2B and 5B parameter variants.

\myparagraph{WAN.}
WAN~\cite{wan2025wan} is a video generation framework available in 1.3B and 14B parameter variants. It uses a spatiotemporal VAE for video compression and a diffusion transformer backbone trained with flow matching. The availability of two model scales allows us to examine scaling behavior within a single architecture.

\subsection{Point Tracking Models Details}
\myparagraph{CoTracker3.}
CoTracker3~\cite{karaev2024cotracker3} uses a ResNet backbone to extract per-frame features and builds a local correlation pyramid as its matching cost. It jointly refines all queried points with an iterative transformer that shares information across tracks. Beyond synthetic data, it is scaled through a semi-supervised protocol that pseudo-labels about 15k real-world videos with off-the-shelf trackers.

\myparagraph{TAP-Net.}
TAP-Net~\cite{doersch2022tap} uses a ResNet backbone and is trained only on the synthetic Kubric dataset. It computes a global cost volume between the query feature and every spatial location in each frame, and then directly regresses point position and occlusion. This global matching closely resembles the zero-shot evaluation protocol used in our analysis.

\myparagraph{TAPIR/BootsTAPIR.}
TAPIR~\cite{doersch2023tapir} combines the global matching of TAP-Net with a PIPs-style~\cite{harley2022particle} iterative refinement stage over a ResNet feature pyramid, and is trained on Kubric. BootsTAPIR~\cite{doersch2024bootstap} extends TAPIR with a bootstrapped self-training stage that additionally leverages large-scale real-world videos, improving robustness on in-the-wild scenes.

\myparagraph{VGGT.}
VGGT~\cite{wang2025vggt} is a feed-forward transformer with full 3D attention built on a DINOv2~\cite{oquab2023dinov2} backbone. Rather than being tracking-specific, it is trained jointly on multiple 3D perception tasks, including camera pose, depth, and point map estimation as well as point tracking, and predicts trajectories through a dedicated tracking head.

\subsection{Evaluation Benchmarks}
\label{supp:eval_protocol}
\myparagraph{TAP-Vid Benchmarks~\cite{doersch2022tap}.} TAP-Vid-DAVIS consists of 30 videos from the DAVIS 2017 validation set~\cite{pont20172017}, while TAP-Vid-Kinetics comprises 1,144 YouTube videos from the Kinetics-700-2020 validation set~\cite{carreira2017quo}. RGB-Stacking is a synthetic benchmark from the TAP-Vid suite, with videos of a simulated arm stacking geometric objects. RoboTAP~\cite{vecerik2024robotap} consists of real-world videos of a robot performing diverse manipulation tasks, probing tracking on real objects that are frequently occluded and re-grasped. Although widely used, these TAP-Vid benchmarks have known limitations for assessing real-world robustness~\cite{demler2025tracker}: most points are static, and tracks rarely require re-identification after occlusion.

\myparagraph{ITTO-MOSE.} Compared to TAP-Vid~\cite{doersch2022tap}, ITTO benchmark~\cite{demler2025tracker} features more dynamic point trajectories and higher track reappearance counts, better reflecting real-world complexity.
Following~\cite{demler2025tracker}, we partition tracks along two dimensions to probe model robustness under varying difficulty levels.
We categorize tracks by their average frame-to-frame displacement relative to the frame diagonal. Three motion levels are defined using thresholds of 0.5\%, 1.5\%, and 5\%: Static ([0\%, 0.5\%)), Moderate ([0.5\%, 1.5\%)), and Fast ([1.5\%, 5\%)). Faster motion poses a greater tracking challenge.
We partition tracks by their reappearance count, \ie, how many times a track becomes visible again after occlusion. Tracks fall into three bins: None ([0, 1)), Occasional ([1, 3)), and Frequent ([3, $\infty$)). More frequent reappearances indicate harder cases that demand robust re-identification across occlusions.


\myparagraph{Corrupted TAP-Vid-DAVIS.} To evaluate robustness to visual artifacts, we apply corruptions from ImageNet-C~\cite{hendrycks2019benchmarking} to TAP-Vid-DAVIS videos following~\cite{nam2023diffusion}. We test at severity level 2 across various corruption types, including noise, blur, and weather. All values reported in Table~\ref{tab:corrupted} are measured by $\delta_{\text{avg}}^x$.

Following standard practice, all videos are resized to $256 \times 256$ for evaluation. During feature extraction, videos are interpolated to each model's native input resolution, and predictions are mapped back to $256 \times 256$ for metric computation.

\subsection{Zero-shot Point Tracking Details}
\label{supp:zeroshot_method}

\subsubsection{Feature extraction.}

For each model, we empirically find the descriptor type (hidden state or query-key projection) and layer that yield the best temporal matching performance. All features are extracted at a unified spatial resolution of $30 \times 52$ for fair comparison.

For DINOv2~\cite{oquab2023dinov2} and DINOv3~\cite{simeoni2025dinov3}, we use the output features of the final layer as both query point and target frame descriptors. For V-JEPA2~\cite{assran2025v}, we use hidden state features from the 24th layer ($l{=}23$). Since V-JEPA2 compresses the temporal dimension by $1/2$ during patch embedding, we repeat each frame twice to obtain one-to-one frame-mapped features with $480 \times 832$ input resolution. For SVD~\cite{blattmann2023stable}, we extract hidden state features from the 3rd upblock layer of the UNet decoder.

For SD3~\cite{esser2024scaling}, we use query and key projection features from the 10th layer ($l{=}9$), extracted at the final denoising timestep. The query projection serves as the query point descriptor and the key projection serves as the target frame descriptor. Using hidden state features instead resulted in substantially lower tracking performance.

For video DiTs~\cite{li2024hunyuan, yang2024cogvideox, wan2025wan}, we encode each frame independently to avoid temporal compression in the VAE encoder, obtaining per-frame latents. We then extract query-key projection features at the final denoising timestep from the layer that yields the best matching performance: $l{=}17$ for CogVideoX-2B and CogVideoX-5B~\cite{yang2024cogvideox}, $l{=}16$ for HunyuanVideo~\cite{li2024hunyuan}, and $l{=}15, 14$ for WAN2.1-1.3B and WAN2.1-14B~\cite{wan2025wan}, respectively.

\subsubsection{Point trajectory estimation.}

Given the extracted features, we compute a matching cost $\mathcal{C}_{1,j}(P_1, p)$ between the query point feature in the first frame and all spatial locations in the $j$-th frame. For video DiTs, following~\cite{nam2025emergent}, we use the scaled dot-product attention score $\text{Softmax}(QK^\top / \sqrt{d})$ to mirror the model's internal matching mechanism. For all other models, we compute the dot product of $\ell_2$-normalized features.

We then find corresponding points $P_j$ in the $j$-th frame for starting point $P_1$ in the first frame via argmax:
\begin{equation}
    P_j
    = \text{Argmax}_{p\in\Omega}
    \, \mathcal{C}_{1,j}(P_1, p),
\end{equation}
where $\Omega$ is the spatial domain. Final point trajectories are obtained by concatenating the starting point with estimated matches, then upscaling to original coordinates through linear interpolation:
\begin{equation}
    \mathcal{T}_{1:F} = \text{Interp}(\text{Concat}(P_1, P_2, \dots, P_F)),
\end{equation}
where $F$ is length of frame sequence.
We report $\delta_{\text{avg}}^x$, the average percentage of correctly tracked points across distance thresholds of 1, 2, 4, 8, and 16 pixels.

\section{Architecture Details}
\label{supp:arch_details}
This section provides the detailed architecture of DiTracker summarized in Section~\ref{sec:method}, including the multi-scale local matching cost, its fusion with the ResNet branch, and the iterative trajectory estimation module adopted from CoTracker3~\cite{karaev2024cotracker3}.

\subsection{Local 4D Matching Cost Computation}
We construct a hierarchical local cost pyramid with varying receptive fields to capture multi-scale query-key relationships. To build a feature pyramid with $S$ scales, we interpolate features to resolution $\tfrac{H}{r\times2^{s-1}} \times \tfrac{W}{r\times2^{s-1}}$, where $r$ denotes the model stride and $s \in \{1,\ldots,S\}$ is the scale factor. For notational simplicity, we omit layer and head indices:
\begin{align}
    Q_i^s &= \text{Interpolate}^{s}(Q_i), \\
    K_i^s &= \text{Interpolate}^{s}(K_i),
\end{align}
where $Q_i^s, K_i^s \in \mathbb{R}^{\tfrac{H}{r\times2^{s-1}} \times \tfrac{W}{r\times2^{s-1}} \times d_\text{head}}$ are interpolated query and key features at scale $s$.
At each scale $s$, we extract local features around points of interest using bilinear sampling. Local query features $q_{i}^s$ are sampled within a $\Delta$-sized neighborhood centered at the query point $\mathbf{p} = (\mathbf{x}, \mathbf{y})$ in the query frame $i$:
\begin{equation}
\label{equation:query_sampling}
    q_{i}^{s} =
    \left[
    Q_{i}^{s}\!\left(\tfrac{\mathbf{x}}{r\times2^{s-1}} + \delta,\; \tfrac{\mathbf{y}}{r\times2^{s-1}} + \delta\right)
    : \|\delta\|_{\infty} \leq \Delta
    \right],
\end{equation}
and local key features $k_j^s$ are sampled around the estimated point $P_j = (x_j, y_j)$ across all frame indices $j \in \{1,\ldots,F\}$:
\begin{equation}
\label{equation:key_sampling}
k_j^{s} =
\left[
K_j^{s}\!\left(\tfrac{x_j}{r\times2^{s-1}} + \delta,\; \tfrac{y_j}{r\times2^{s-1}} + \delta\right)
: \|\delta\|_{\infty} \leq \Delta
\right],
\end{equation}
where $\delta \in \mathbb{Z}$ and $q_i^s, k_j^s \in \mathbb{R}^{d_\text{head} \times (2\Delta + 1)^2}$.
We then construct a \textit{local 4D matching cost} $\mathcal{C}_{i,j}^{s,\text{DiT}}$ between local features of the query frame $i$ and target frame $j$ with a softmax operation:
\begin{equation}
\label{equation:local_supp}
\mathcal{C}_{i,j}^{s, \text{DiT}} = \text{Softmax} \!\left(\frac{q_{i}^s (k_j^s)^\top}{\sqrt{d_\text{head}}}\!\right) \in \mathbb{R}^{(2\Delta + 1)^4}.
\end{equation}

\subsection{Multi-scale Cost Fusion}
We extract ResNet features $\Phi_i = \text{ResNet}(X_i)$, construct a multi-scale feature pyramid, and sample local features $\varphi_{i}^s$ and $\varphi_{j}^s$ around the query point $\mathbf{p}$ in frame $i$ and around the estimated point $P_j$ in frame $j$, respectively. The local ResNet matching cost is:
\begin{equation}
\mathcal{C}_{i,j}^{s,\text{ResNet}} = \text{Softmax} \!\left(\frac{\varphi_{i}^s (\varphi_j^s)^\top}{\sqrt{d_\text{ResNet}}}\!\right) \in \mathbb{R}^{(2\Delta + 1)^4},
\end{equation}
where $d_\text{ResNet}$ is the channel dimension of ResNet features. We flatten and concatenate the DiT and ResNet matching costs at each scale to obtain the fused cost:
\begin{equation}
\label{equation:fusing_supp}
\mathcal{C}_{i,j}^{s,\text{fused}} = \left[\text{Flatten}(\mathcal{C}_{i,j}^{s,\text{DiT}}), \text{Flatten}(\mathcal{C}_{i,j}^{s,\text{ResNet}})\right] \in \mathbb{R}^{2(2\Delta + 1)^4}.
\end{equation}
Finally, we project costs from all scales to cost embeddings $E_{j}$ using an MLP:
\begin{equation}
E_{j} = \text{MLP}(\mathcal{C}_{i,j}^{1,\text{fused}}, \ldots, \mathcal{C}_{i,j}^{S,\text{fused}}) \in \mathbb{R}^{d_E},
\end{equation}
where $d_E$ is the channel dimension of cost embeddings.

\subsection{Trajectory Estimation}
We estimate point trajectories using an iterative transformer-based refinement module following~\cite{karaev2024cotracker3}.
We first initialize trajectories by broadcasting the query point location to all frames with zero visibility and confidence. At each iteration $t$, we then calculate per-frame displacement embeddings $\eta_{i \to i+1} = \eta(P_{i+1}-P_i)$ using Fourier Encoding $\eta$, then construct input tokens $G_i$ by concatenating displacement embeddings $\eta_{i-1\to i}$ and $\eta_{i\to i+1}$, visibility $V_i$, confidence $C_i$, and cost embeddings $E_i$ for every query point $\mathbf{p}$. A tracking head $\Psi$ with factorized temporal and query-point attention processes these tokens to predict residual updates: $P^{(t+1)} = P^{(t)} + \Delta P^{(t)}$, $V^{(t+1)} = V^{(t)} + \Delta V^{(t)}$, $C^{(t+1)} = C^{(t)} + \Delta C^{(t)}$, where ${\Delta P^{(t)}, \Delta V^{(t)}, \Delta C^{(t)} = \Psi(G)}$.
After each update, we resample correlation features at the refined locations. After $T$ iterations, we obtain the final trajectory predictions $\{P_i, V_i, C_i\}_{i=1}^F$.

\subsection{Training Loss}
\label{supp:training_loss}
Following ~\cite{karaev2024cotracker3}, our training objective consists of three complementary loss terms that supervise trajectory estimation, visibility, and confidence prediction across all refinement iterations as $\mathcal{L} = \mathcal{L}_{\text{track}} + \mathcal{L}_{\text{vis}} + \mathcal{L}_{\text{conf}}$.

\paragraph{Trajectory supervision.} We supervise coordinate predictions for both visible and occluded points using the Huber loss $\text{Huber}(\cdot)$ with a 6-pixel threshold. To encourage the model to focus on accurately tracking visible points, we apply asymmetric weighting based on occlusion state:
\begin{equation}
    \mathcal{L}_\text{track} = \sum_{t=1}^T \gamma^{T-t} \left(\frac{\mathbb{1}_\text{occ}}{5} + \mathbb{1}_\text{vis} \right) \cdot \text{Huber}(P^{(t)}, \hat{P}),
\end{equation}
where $P^{(t)}$ denotes predicted coordinates at iteration $t \in \{1,\ldots,T\}$, $\hat{P}$ is the ground truth, $\mathbb{1}_{\text{occ}} \in \{0,1\}$ and $\mathbb{1}_{\text{vis}} \in \{0,1\}$ are binary visibility states indicating whether the point is occluded or visible, and $\gamma = 0.8$ is a temporal discount factor that assigns higher importance to later refinement stages. This weighting scheme assigns $5\times$ lower weight to occluded points, prioritizing accurate tracking of visible points.

\paragraph{Visibility supervision.} Visibility predictions are supervised with Binary Cross Entropy (BCE) loss $\text{CE}(\cdot)$ using ground truth visibility annotations:
\begin{equation}
\mathcal{L}_\text{vis} = \sum_{t=1}^T \gamma^{T-t} \text{CE}(\sigma(V^{(t)}), \hat{V}), 
\end{equation}
where $V^{(t)}$ and $\hat{V}$ denote predicted and ground truth visibility respectively, and $\sigma(\cdot)$ is the sigmoid function applied to visibility logits $V^{(t)}$.

\paragraph{Confidence supervision.} We supervise confidence scores using BCE loss. For each iteration, the ground truth confidence label is defined by whether the current prediction falls within 12 pixels of the ground truth location:
\begin{equation}
\mathcal{L}_\text{conf} = \sum_{t=1}^T \gamma^{T-t} \text{CE}\big(\sigma(C^{(t)}), \mathbb{1}[\|P^{(t)} - \hat{P}\|_2 < 12]\big),
\end{equation}
where $\sigma(\cdot)$ is the sigmoid function applied to predicted confidence logits $C^{(t)}$, and $\mathbb{1}[\cdot]$ is the indicator function that returns 1 when the condition holds and 0 otherwise.

\begin{figure*}[!t]
    \centering
    \includegraphics[width=\linewidth]{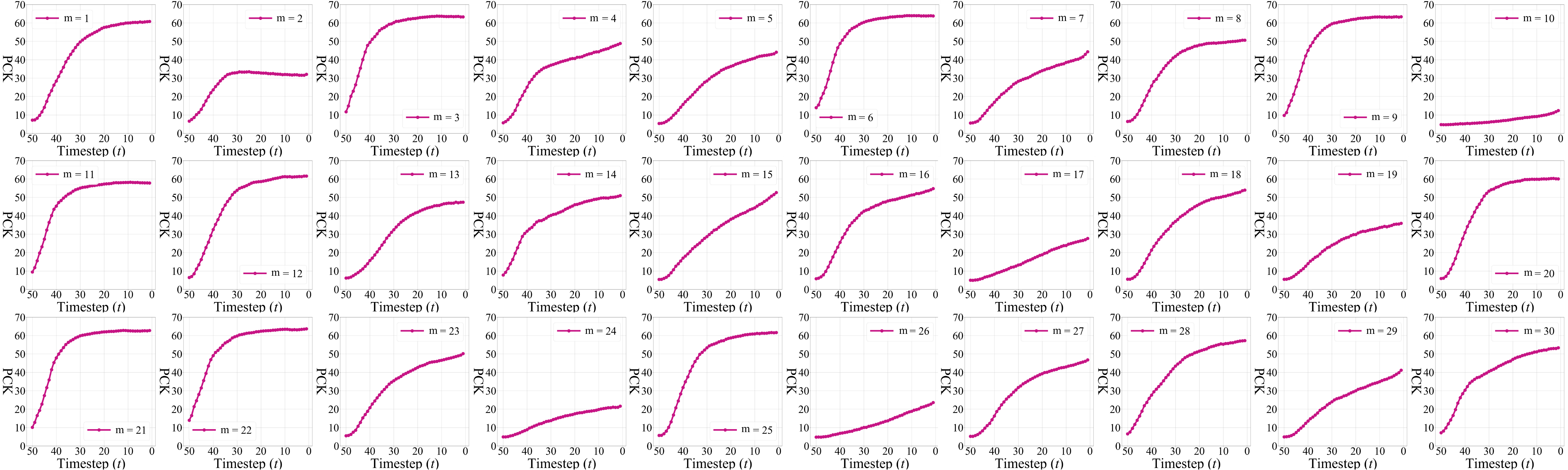}
    \vspace{-15pt}
    \caption{\textbf{Attention head analysis in 18th layer of CogVideoX-2B~\cite{yang2024cogvideox}.} Using a subset of heads with high accuracy instead of concatenating all heads significantly reduces memory requirements while maintaining strong performance.
    }
    \label{fig:supp_head_analysis}
\end{figure*}
\begin{table*}[!t]
\centering
\caption{\textbf{Computation cost.} We compute cost on a single RTX A6000 GPU. Inference cost is computed using 8 frames and training cost is computed using 1 batch.}
\label{tab:cost}
\resizebox{\linewidth}{!}{
\begin{tabular}{l|cccccc|cc||cc}
\toprule
\multirow{3}{*}{Method} 
& \multicolumn{8}{c||}{\textbf{Inference Cost}} 
& \multicolumn{2}{c}{\textbf{Training Cost}} \\

& \multicolumn{6}{c|}{Time (s)} 
& \multirow{2}{*}{\shortstack{Throughput\\(points/sec)}} 
& \multirow{2}{*}{\shortstack{FLOPs\\per point (G)}} 
& \multirow{2}{*}{\shortstack{VRAM\\(GB)}} 
& \multirow{2}{*}{\shortstack{Wall-clock\\(ms/step)}} \\

& $10^0$ point & $10^1$ points & $10^2$ points & $10^3$ points & $10^4$ points & $10^5$ points 
& & & & \\
\midrule

CoTracker3~\cite{karaev2024cotracker3} 
& 0.08 & 0.08 & 0.08 & 0.26 & 2.44 & OOM 
& 4100.71 & 0.70 
& 20.68 & 881.06 \\

\textbf{DiTracker} 
& 2.42 & 2.43 & 2.45 & 2.53 & 3.69 & OOM 
& 2707.04 & 1.28 
& 36.18 & 9479.88 \\

\bottomrule
\end{tabular}}
\end{table*}

\section{Implementation Details}
\label{supp:method_imde}
We train DiTracker for 39k iterations on 4 NVIDIA RTX A6000 GPUs. We use an input resolution of $480{\times}720$, chosen based on CogVideoX-2B's native input resolution. The tracking head uses the following hyperparameters: feature pyramid with $S{=}4$ scales, local correlation radius $\Delta{=}3$, model stride $r{=}4$, LoRA rank 128, and channel dimension $d{=}64$. We set the maximum sequence length to 46 frames with a batch size of 4. These are modest compared to the larger batch sizes and longer sequences used in prior work, yet video DiT features enable state-of-the-art performance even under these limited training conditions.

We extract features from the 18th layer ($l{=}18$) of CogVideoX-2B~\cite{yang2024cogvideox}. To select the optimal attention heads, we analyze zero-shot point tracking performance across all denoising timesteps and all heads in this layer. \Cref{fig:supp_head_analysis} shows that heads $m{=}3,6,9,21$ achieve high accuracy at final denoising timestep. Using a subset of heads instead of concatenating all heads significantly reduces memory requirements while maintaining strong performance.

\section{Additional Qualitative Results}
\label{supp:visualize}
\begin{figure*}[t]
    \centering
    \includegraphics[width=1\linewidth]{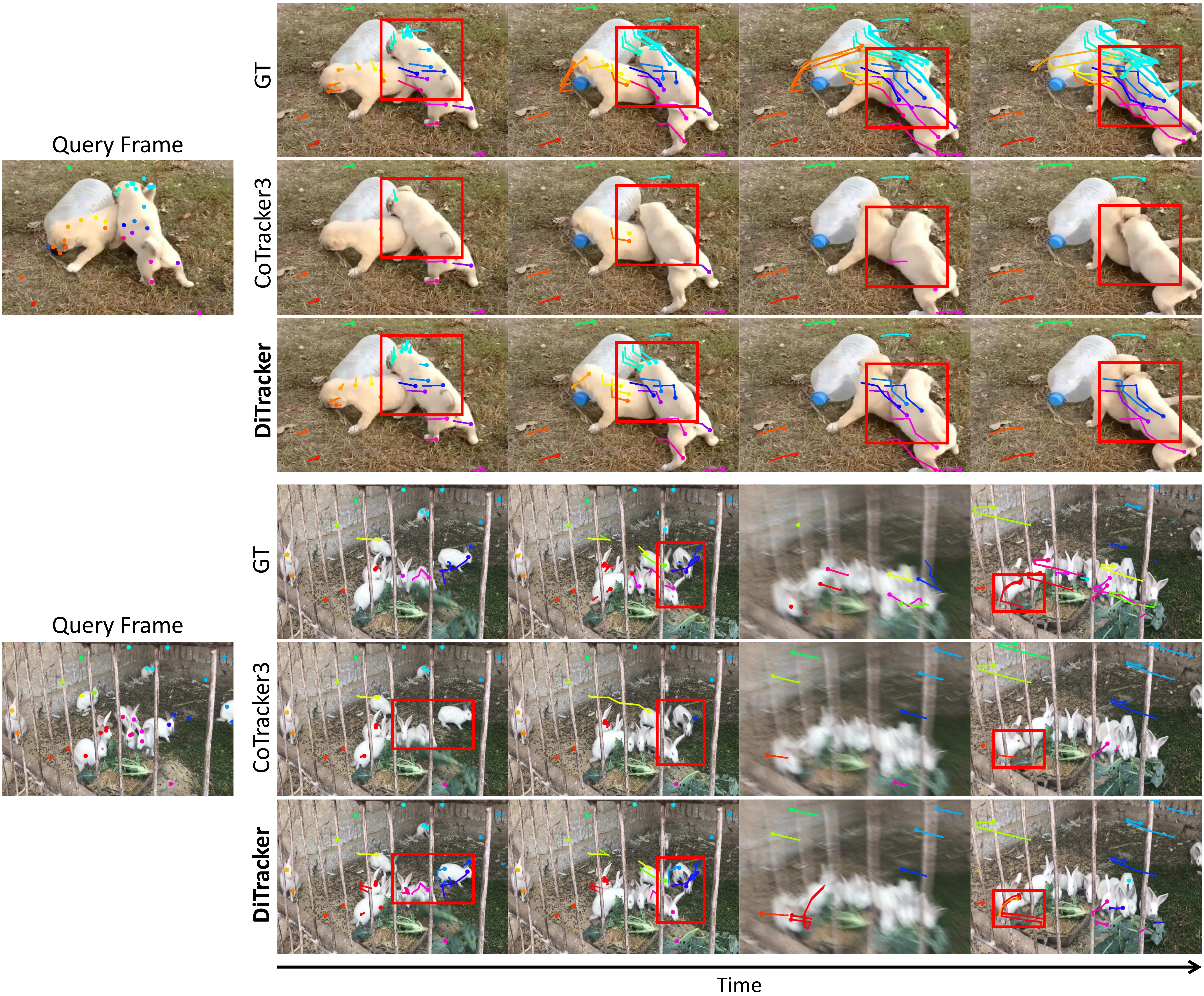}
    \vspace{-15pt}
   \caption{\textbf{Qualitative results on ITTO-MOSE~\cite{demler2025tracker} benchmark.} Our DiTracker predicts more smooth and accurate point trajectories under challenging real-world scenarios, including large displacement, motion blur, and occlusions, even surpassing CoTracker3~\cite{karaev2024cotracker3}.}
    \label{fig:supp_itto_qual_1}
\end{figure*}

\begin{figure*}[t]
    \centering
    \includegraphics[width=1\linewidth]{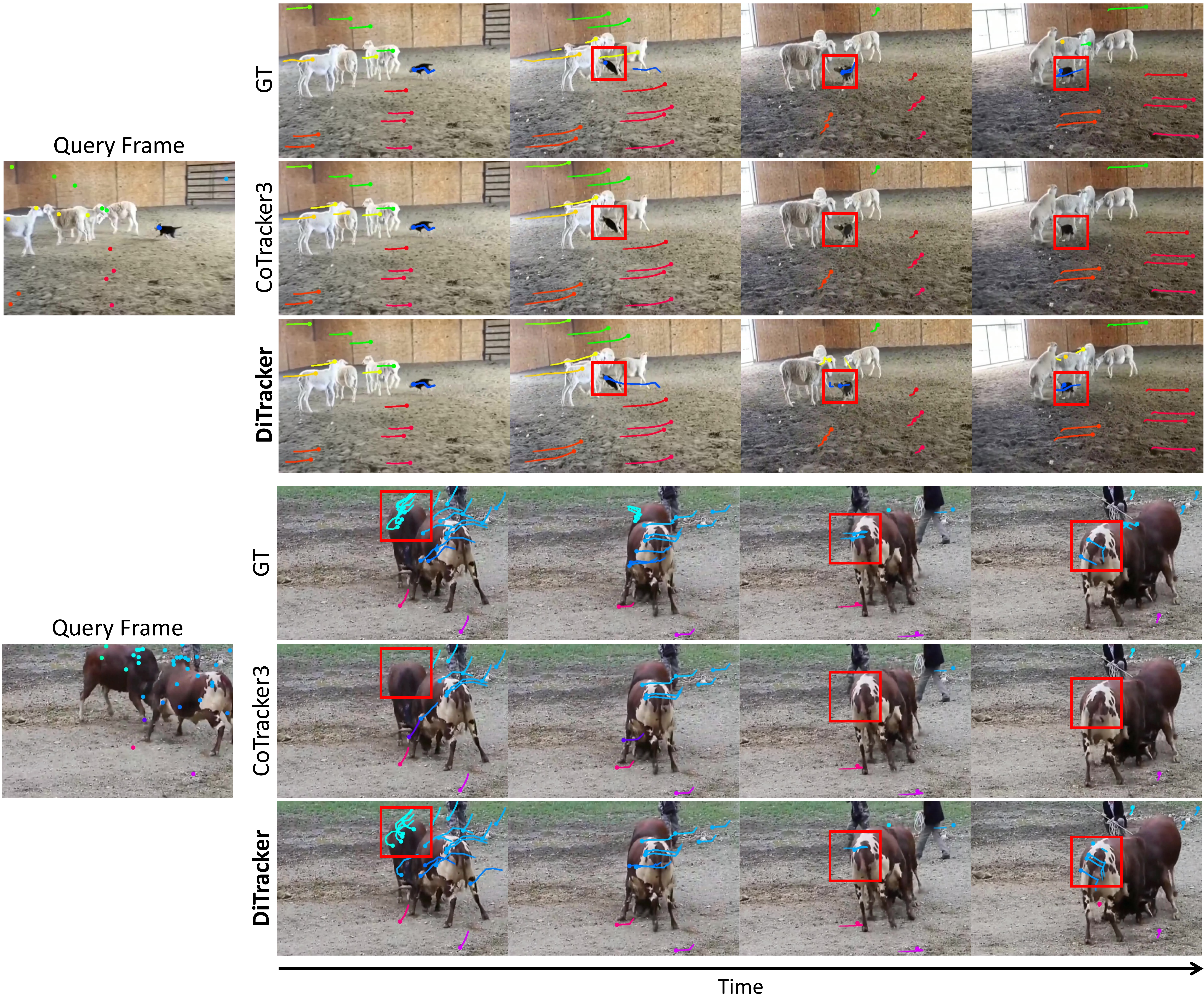}
    \vspace{-15pt}
   \caption{\textbf{Qualitative results on ITTO-MOSE~\cite{demler2025tracker} benchmark.} Our DiTracker predicts more smooth and accurate point trajectories under challenging real-world scenarios, including large displacement, motion blur, and occlusions, even surpassing CoTracker3~\cite{karaev2024cotracker3}.}
    \label{fig:supp_itto_qual_2}
\end{figure*}

\begin{figure*}[h]
    \centering
    \includegraphics[width=\linewidth]{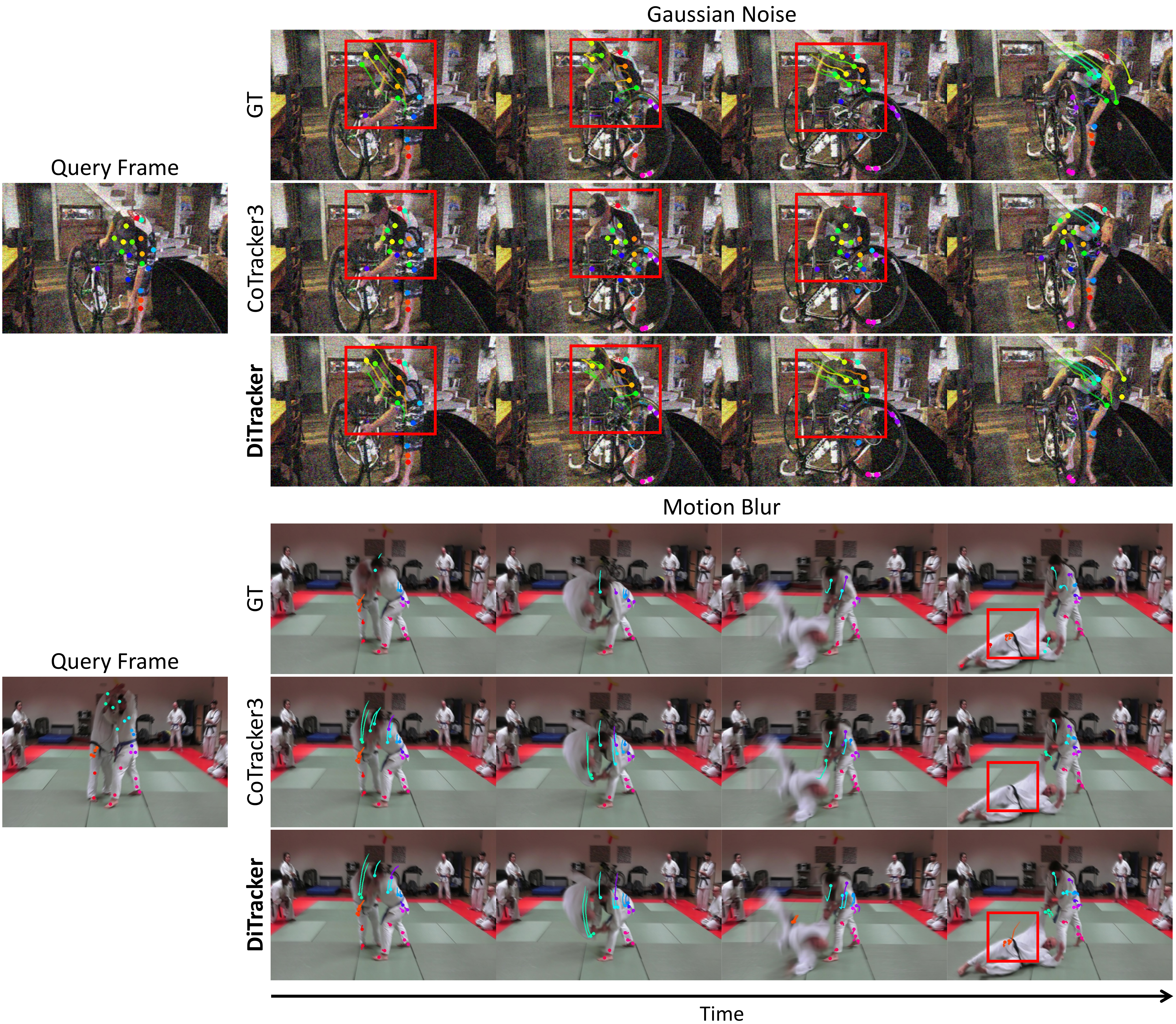}
    \caption{\textbf{Qualitative results on TAP-Vid-DAVIS~\cite{doersch2022tap} with corruptions from ImageNet-C~\cite{hendrycks2019benchmarking}.} Our DiTracker predicts more accurate point trajectories under gaussian noise and motion blur corruptions compared to CoTracker3~\cite{karaev2024cotracker3}.}
    \label{fig:supp_corruption_qual_3}
\end{figure*}

\begin{figure*}[h]
    \centering
    \includegraphics[width=\linewidth]{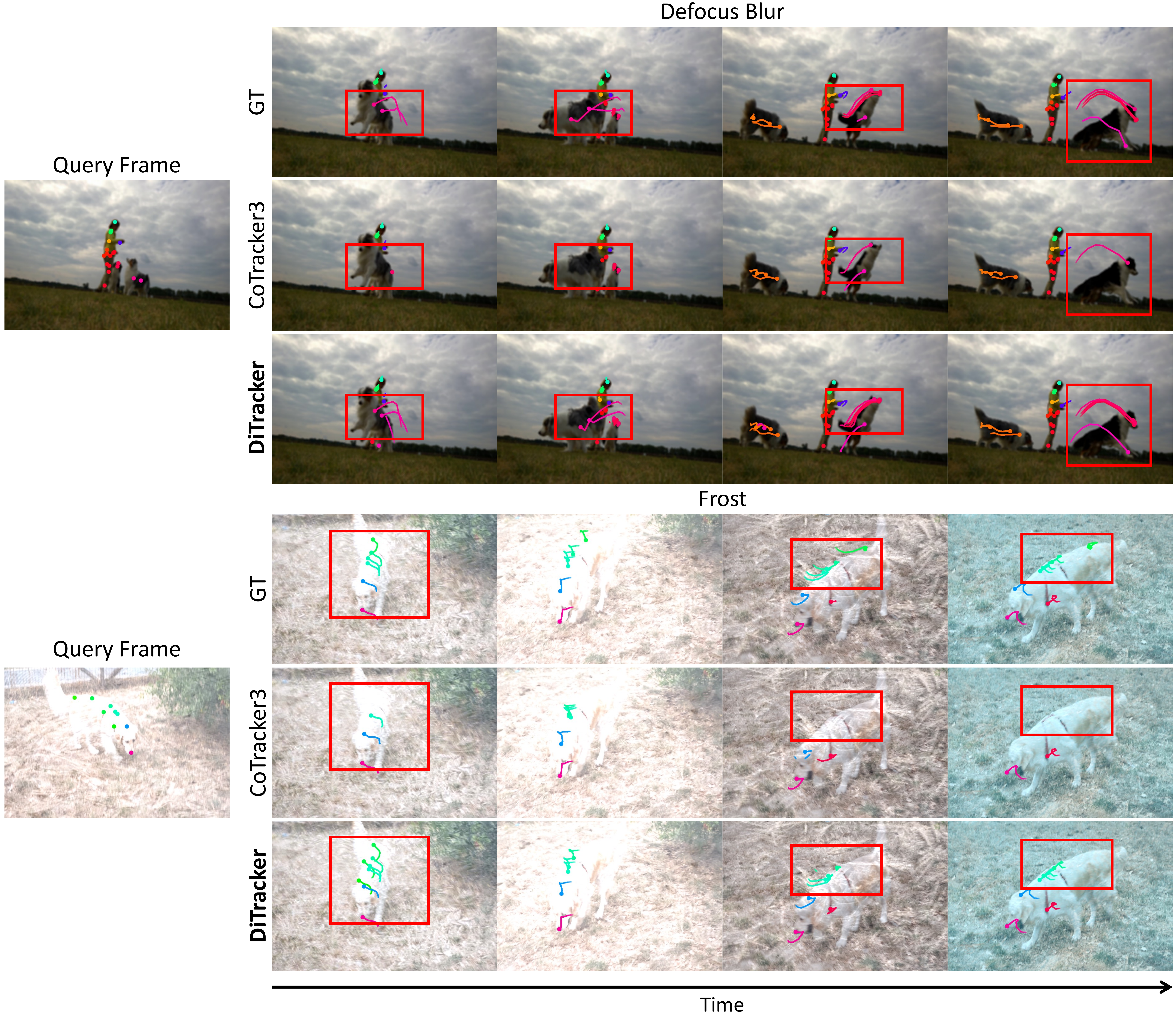}
    \caption{\textbf{Qualitative results on TAP-Vid-DAVIS~\cite{doersch2022tap} with corruptions from ImageNet-C~\cite{hendrycks2019benchmarking}.} Our DiTracker predicts more accurate point trajectories under defocus blur and frost corruptions compared to CoTracker3~\cite{karaev2024cotracker3}.}
    \label{fig:supp_corruption_qual_1}
\end{figure*}

\begin{figure*}[h]
    \centering
    \includegraphics[width=\linewidth]{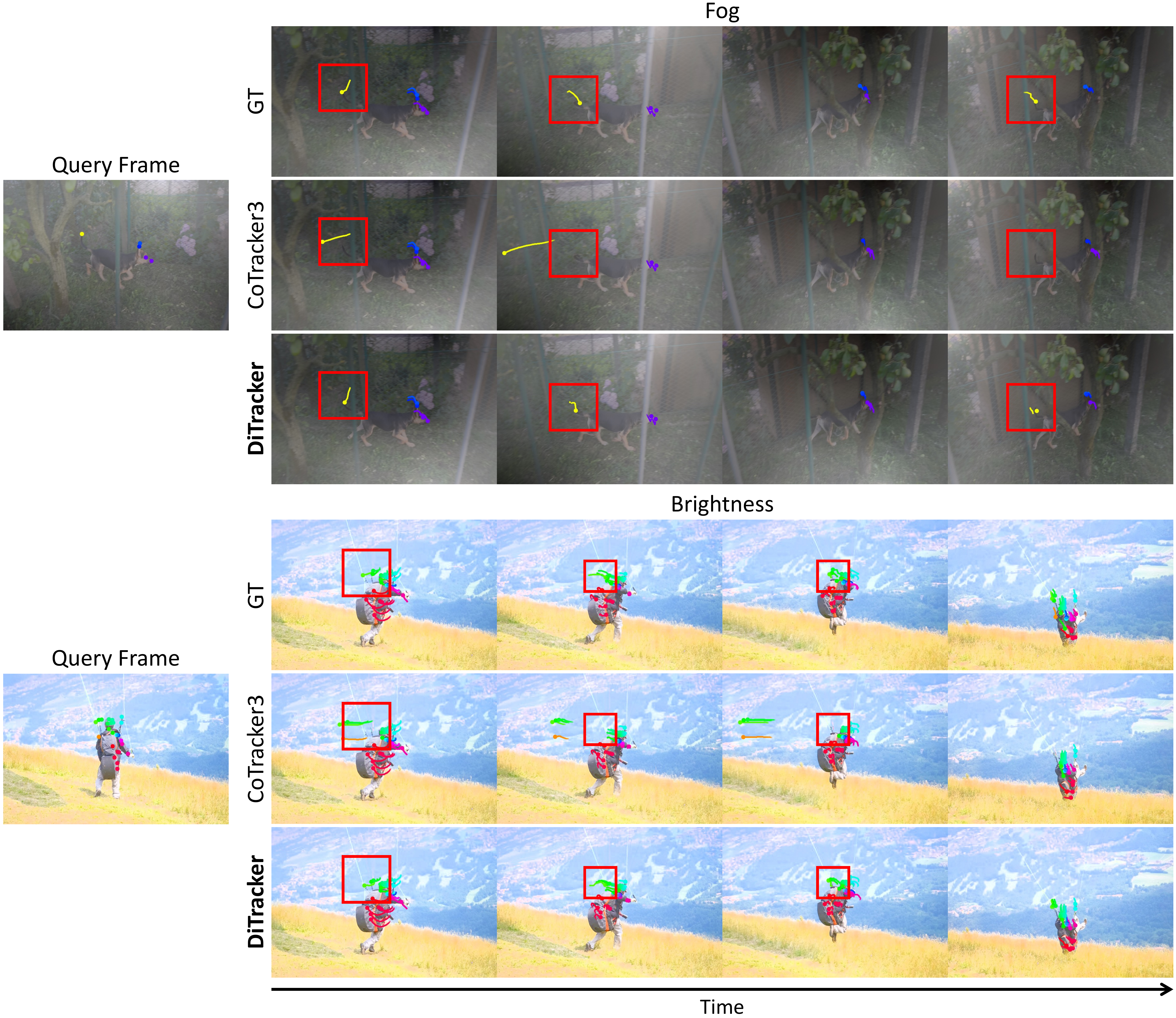}
    \caption{\textbf{Qualitative results on TAP-Vid-DAVIS~\cite{doersch2022tap} with corruptions from ImageNet-C~\cite{hendrycks2019benchmarking}.} Our DiTracker predicts more accurate point trajectories under fog and brightness corruptions compared to CoTracker3~\cite{karaev2024cotracker3}.}
    \label{fig:supp_corruption_qual_2}
\end{figure*}

\Cref{fig:supp_itto_qual_1}, \ref{fig:supp_itto_qual_2}, \ref{fig:supp_corruption_qual_3}, \ref{fig:supp_corruption_qual_1}, and \ref{fig:supp_corruption_qual_2} shows visualizations on ITTO-MOSE and TAP-Vid-DAVIS with corruptions from ImageNet-C~\cite{hendrycks2019benchmarking}. Our DiTracker produces more accurate and stable trajectories across both benchmarks, demonstrating superior robustness than CoTracker3~\cite{karaev2024cotracker3}.


\section{Future work}
\label{supp:limitation}
While our work demonstrates the effectiveness of video DiT features for point tracking, opportunities for improvement remain that we leave for future work. As shown in \Cref{tab:cost}, at inference time, DiTracker requires more time and memory than CoTracker3 due to the cost of extracting features from the large video diffusion model.

\clearpage
\newpage

\end{document}